\documentclass{ieeeaccess}
\usepackage{cite}
\usepackage{amsmath,amssymb,amsfonts}
\usepackage{graphicx}
\usepackage{textcomp}

\usepackage[shortlabels]{enumitem}
\usepackage{bbm}
\usepackage{indentfirst}
\usepackage{xcolor}
\usepackage{booktabs}
\usepackage{multirow}
\usepackage{pbox}
\usepackage{float}

\usepackage{tabularx,ragged2e}

\usepackage{balance}
\usepackage{algorithm}
\usepackage{algpseudocode}

\usepackage{float}
\usepackage{subfigure}

\usepackage{url}

\usepackage{stfloats}

\usepackage{bm}
\makeatletter
\AtBeginDocument{\DeclareMathVersion{bold}
\SetSymbolFont{operators}{bold}{T1}{times}{b}{n}
\SetSymbolFont{NewLetters}{bold}{T1}{times}{b}{it}
\SetMathAlphabet{\mathrm}{bold}{T1}{times}{b}{n}
\SetMathAlphabet{\mathit}{bold}{T1}{times}{b}{it}
\SetMathAlphabet{\mathbf}{bold}{T1}{times}{b}{n}
\SetMathAlphabet{\mathtt}{bold}{OT1}{pcr}{b}{n}
\SetSymbolFont{symbols}{bold}{OMS}{cmsy}{b}{n}
\renewcommand\boldmath{\@nomath\boldmath\mathversion{bold}}}
\makeatother

\def\BibTeX{{\rm B\kern-.05em{\sc i\kern-.025em b}\kern-.08em
    T\kern-.1667em\lower.7ex\hbox{E}\kern-.125emX}}

\begin{document}

\history{Received 22 August 2024, accepted 21 September 2024, date of publication 27 September 2024, \\
	date of current version 21 October 2024.}
\doi{10.1109/ACCESS.2024.3469163}

\title{One Stone, Four Birds: A Comprehensive Solution for QA System Using Supervised Contrastive Learning}
\author{\uppercase{Bo Wang}\authorrefmark{1},
\uppercase{Tsunenori Mine}\authorrefmark{2}, \IEEEmembership{Member, IEEE}}

\address[1]{Graduate School of Information Science and Electrical Engineering, Kyushu University, Fukuoka, 8190395, JAPAN (e-mail: wangbo.rw@gmail.com)}
\address[2]{Faculty of Information Science and Electrical Engineering, Kyushu University, Fukuoka, 8190395, JAPAN (e-mail: mine@ait.kyushu-u.ac.jp)}
\tfootnote{This work was supported in part by JST SPRING No. JPMJSP2136 and JSPS KAKENHI under Grant JP19KK0257, JP21H00907 and JP23H03511}

\markboth
{Bo Wang \headeretal: A Comprehensive Solution for QA System}
{Bo Wang \headeretal: A Comprehensive Solution for QA System}

\corresp{Corresponding author: Bo Wang (e-mail: wangbo.rw@gmail.com).}

\begin{abstract}
This paper presents a novel and comprehensive solution to enhance both the robustness and efficiency of question answering (QA) systems through supervised contrastive learning (SCL). 
Training a high-performance QA system has become straightforward with pre-trained language models, requiring only a small amount of data and simple fine-tuning. 
However, despite recent advances, existing QA systems still exhibit significant deficiencies in functionality and training efficiency. 
We address the functionality issue by defining four key tasks: user input intent classification, out-of-domain input detection, new intent discovery, and continual learning. 
We then leverage a unified SCL-based representation learning method to efficiently build an intra-class compact and inter-class scattered feature space, facilitating both known intent classification and unknown intent detection and discovery. 
Consequently, with minimal additional tuning on downstream tasks, our approach significantly improves model efficiency and achieves new state-of-the-art performance across all tasks.

\end{abstract}


\titlepgskip=-21pt

\maketitle

%\linenumbers
\section{Introduction}\label{sec:intro}

\begin{figure*}[tb]
	\centering
	\includegraphics[width=0.95\linewidth]{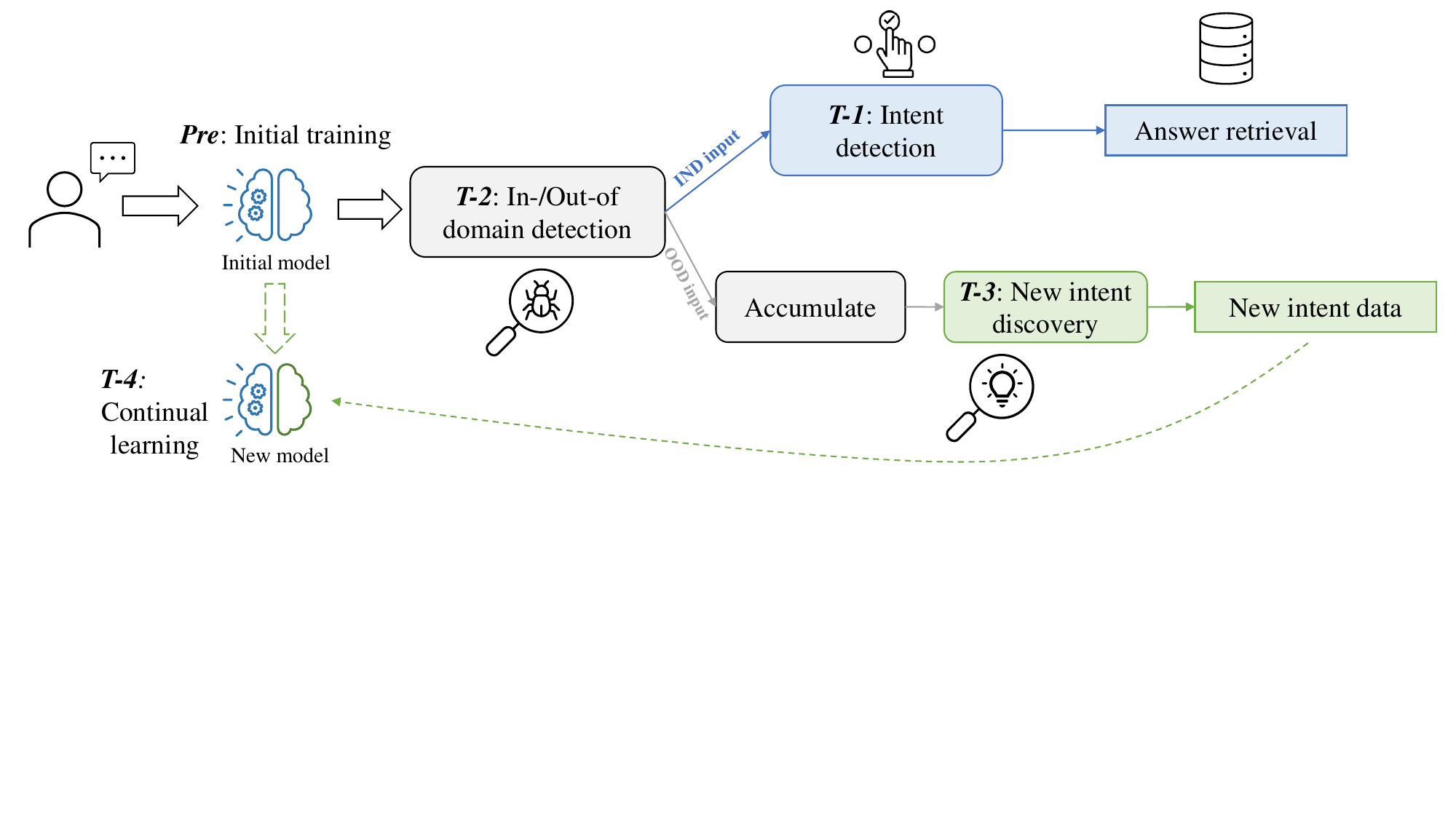}
	\caption{The workflow of the proposed comprehensive and adaptive QA system. First, an initial model is trained using current known data (\textbf{Pre}). Then, during daily usage, an inquiry will first be judged to be an answerable input or not (\textbf{T-2}): if so, proceed to the intent detection (\textbf{T-1}) and retrieve the answer; if not, the unknown inquiry will be temporally saved and be clustered later (\textbf{T-3}). Finally, the initial model is re-trained with newly discovered data to obtain the classification ability to new intents (\textbf{T-4}).}
	\label{fig:workflow}
\end{figure*} 

\PARstart{Q}{uestion} answering (QA) system is a well-established and important application widely used in areas such as healthcare \cite{alzubi2023cobert}, finance \cite{zhu-etal-2021-tat} and e-commerce \cite{gao2021meaningful}.
Despite the world gradually moving into the era of large language models (LLMs), classic QA systems still maintain a strong presence due to their low cost, ease of deployment, and high accuracy in answering questions.

A typical QA system mainly comprises two components: user intent classification and answer retrieval. The efficiency of the former is crucial, as a correct response often depends on accurately determining the user's intent. 
To achieve robust classification, it is essential to efficiently extract features from user input, providing intent-specific and discriminative text representations for classification. 
Initially, text features were extracted manually using techniques like term frequency-inverse document frequency (TF-IDF \cite{robertson2004understanding}) or bag-of-words (BoW \cite{mikolov2013efficient}). With the advent of neural networks, feature extraction became an automated process \cite{pennington-etal-2014-glove,bojanowski2017enriching}. 
Today, the power of pre-trained Transformers like BERT \cite{devlin2018bert} has made it much easier to extract features and build classification models, requiring only a small amount of data with Cross-Entropy (CE) loss fine-tuning \cite{qu2019bert,do2022developing}.

It seems that the current QA system building is simple and mature, with nearly nothing to be changed.
However, from the perspective of building a comprehensive system, current QA systems still face unresolved issues at both internal and external levels, which have not been fully addressed yet:

\begin{enumerate}[$\bullet$]
	\item \textit{\textbf{Externally, or at the system level}}: Current QA systems typically adopt a ``training-as-complete'' policy, meaning that once the model has finished training, its knowledge is not updated. This approach leaves the system vulnerable to unseen, unknown inputs (as out-of-domain, OOD inputs \cite{fei-liu-2016-breaking}). To maintain the robustness and security of the model, there should be mechanisms to handle these inputs. Furthermore, even when new data are collected to upgrade the system, the common practice is to re-train a new model from scratch, leading to a waste of labor and resources.
	\item \textit{\textbf{Internally, or in training efficiency}}: Increasing studies indicate that the BERT+CE training combination still has efficiency problems. For BERT, the raw text feature distribution is anisotropic \cite{wang2020improving}. Additionally, when using CE as the fine-tuning loss, text features of different intents often remain mixed with unclear boundaries in the embedding space \cite{oord2018representation}. Both issues hinder training efficiency and model performance.
\end{enumerate}

In this paper, aiming to build a more robust QA system, we propose a novel and comprehensive solution to address both issues. 
To tackle the systematic issue, including the basic intent classification function, we define four key tasks and design a workflow between them (as shown in Figure \ref{fig:workflow}), enabling the system to be self-consistent and recurrently updatable. Specifically, the tasks are:

\begin{enumerate}[T-1]
	\item \textit{\textbf{User input intent classification}}: The fundamental task for a QA system.
	\item \textit{\textbf{OOD input detection}}: Detect abnormal inputs to prevent misleading output. The detected OOD texts are saved as source data for Task 3.
	\item \textit{\textbf{New intent discovery}}: Utilize detected OOD texts to discover new possible intents once they accumulate to a sufficient quantity.
	\item \textit{\textbf{Continual learning}}: Use newly discovered and pseudo-labeled data to incrementally train the initial classification model, enhancing the system's capability to handle the latest inquiries.
\end{enumerate}

To address the training efficiency problem, we leverage supervised contrastive learning (SCL, \cite{khosla2020supervised}) to replace CE fine-tuning as the unified text representation optimization method, which requires minimal tuning to achieve all four task objectives. 
We choose SCL because, in our previous research on OOD detection \cite{wang2023optimizing}, it demonstrated excellent capability in upstream feature optimization, allowing an accurate OOD text detection using only a simple distance-based algorithm in downstream.

\begin{figure*}[tb]
	\centering
	\includegraphics[width=0.9\linewidth]{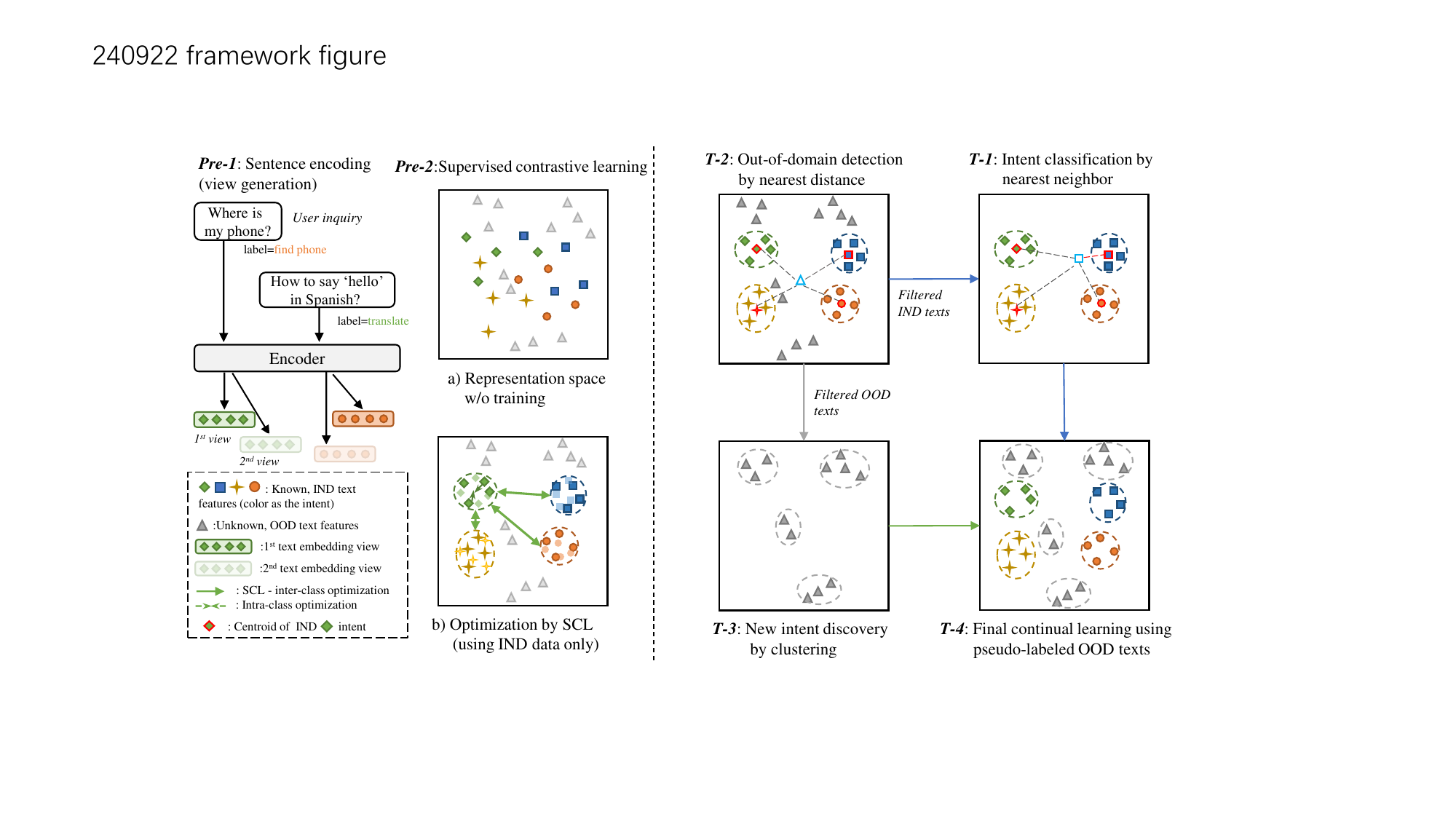}
	\caption{Diagrams of SCL's representation optimization procedure, and solution for each task. Left: the training and optimization effect by SCL, which gathers same-intent text features together and pushes different features further apart, providing an underlying embedding space for following tasks. Right: the classification/detection processes of T-1 to T-4 after SCL. (Together with Figure \ref{fig:workflow} for better understanding)}
	\label{fig:framework}

\end{figure*}

More specifically, the advantage of SCL over CE fine-tuning lies in its ability to enhance both learning efficiency and feature separation. 
As shown in Figure \ref{fig:framework}-b, SCL employs a simple but clear optimization target: pulling same-label text features together and pushing different ones further apart, resulting in an intra-class compact and inter-class scattered, well-organized embedding space.
This clear separation between clusters facilitates easier in-domain (IND) intent classification and IND/OOD separation and detection (Figure \ref{fig:framework}, T-1/2). 
Additionally, this structure also supports the aggregation of currently unknown but semantically similar features, creating a conducive environment for new intent discovery and continual learning (Figure \ref{fig:framework}, T-3/4).
Consequently, even with the simplest algorithms in downstream tuning, all task objectives can be achieved effortlessly and efficiently.

Our contributions can be summarized as follows:
\begin{enumerate}
	\item For the first time, we clearly identify the systematic and efficiency issues in current QA systems and propose a comprehensive solution involving four key tasks in a natural workflow to enhance the training process, resulting in a robust and comprehensive system.
	\item To achieve the task objectives, we leverage SCL as the unified representation learning approach, which requires minimal tuning in downstream tasks.
	\item Our experiments demonstrate that, despite its conciseness and simplicity, the proposed solution significantly improves global feature optimization efficiency and outperforms state-of-the-art methods across all tasks, complementing the deficiencies of current QA systems.\footnote{Code is available at \url{https://github.com/bowang-rw-02/scl-one-for-four}}
\end{enumerate}

\section{Related Works}\label{sec:related}
\subsection{Representation learning and intent classification model building in QA system}
As introduced in Section \ref{sec:intro}, the efficiency of a QA system heavily depends on the performance of the intent classification module, where the representation learning (RepL) process plays a key role in it. Here, a good RepL method can capture the intrinsic patterns and structures of the data, and thus providing a solid foundation for downstream classification and other tasks.
Early RepL research focused on manually extracting meaningful features from raw data, with preliminary methods including TF-IDF \cite{robertson2004understanding} and BoW \cite{mikolov2013efficient}. Subsequently, the development of word embeddings extracted via neural networks automated the feature extraction process \cite{pennington-etal-2014-glove,bojanowski2017enriching}. More recently, with the advent of large-scale language encoders like BERT \cite{devlin2018bert} that are pre-trained on extensive corpora, obtaining high-quality text embeddings has become much easier. The common classification model building approach is to use BERT as the base encoder, with CE fine-tuning as the RepL method, which is widely used in various recent QA systems \cite{wang2021legal,do2022developing,alzubi2023cobert,zheng2024pal}.

\noindent \textit{\textbf{Drawback of BERT+CE mode}}

However, this BERT+CE paradigm is proven to be suboptimal for feature optimization these years. 
Firstly, from a global perspective, \cite{gao2019representation,wang2020improving} reported that the overall text embedding distribution generated by BERT is anisotropic. This means that most embeddings are spatially distributed in an elongated cone-like space rather than uniformly in a sphere, leading to high similarity between any two sentence representations and thus posing challenges for downstream tasks.
Secondly, from a local perspective, CE fine-tuning usually results in a ``loose'' feature arrangement \cite{oord2018representation,khosla2020supervised}, where the boundaries between different intent clusters are barely distinguishable. This not only reduces the performance of known, IND intent classification, but also makes the model prone to confusing IND inputs with future unseen but similar OOD inputs.

\noindent \textit{\textbf{Contrastive learning as a RepL method}}

To overcome above limitations and further improve learning efficiency, the contrastive learning (ConL), has been actively studied recently.
Earliest ConL can be traced back to supervised metric learning (MeL) \cite{sohn2016improved} used in image classification tasks. Different with the vague and loose optimization goal of CE fine-tuning mode, MeL applies a simpler but straightforward target of ``gathering similar examples together and pushing dissimilar ones further apart''. This helps generate compact feature clusters efficiently, and established the foundation of ConL. 
Subsequently, unsupervised ConL (UConL) \cite{chen2020simple,he2020momentum} was proposed, sharing the same strategy but aiming to improve the overall representation uniformity. 
Finally, SCL \cite{khosla2020supervised} incorporates the advantages of both MeL and UConL, thus having the potential to address both global and local feature optimization issues.

Despite these advantages, currently ConL is still mainly examined in computer vision (CV) field. In natural language processing (NLP) field, there are only limited applications, primarily in natural language inference (NLI) tasks using UConL methods \cite{gao-etal-2021-simcse,chuang-etal-2022-diffcse}.
A boarder exploration of SCL's learning effect across various tasks, including its application in QA systems, remains unaddressed, which is exactly our main research target this time.

\subsection{Past work on individual tasks}
To provide a boarder perspective, we briefly review previous methods for addressing the three newly introduced tasks.

\noindent \textit{\textbf{T-2 Out-of-Domain, OOD detection}}: 
Although QA systems have evolved significantly in recent years, they can only work correctly on the known inquires and lack the ability to recognize and reject unknown inputs, which raises the issue of OOD detection \cite{fei-liu-2016-breaking}.
To solve this problem, early methods built a binary IND/OOD detection system using numerous OOD data \cite{hendrycks2018deep}. However, the scope and variety of OOD data can be infinite, making this method impractical. 
Subsequently, more approaches have focused on training networks using only IND data to generate heterogeneous representations \cite{shen-etal-2021-enhancing,colombo2022beyond}, then applied various downstream algorithms, such as local outlier factor (LOF) \cite{zhou-etal-2023-two}, Gaussian discriminant analysis (GDA) \cite{xu-etal-2020-deep}, energy score \cite{liu2020energy,ouyang-etal-2021-energy}, and Mahalanobis distance \cite{podolskiy2021revisiting,lin-gu-2023-flats}, to distinguish IND and OOD text features. 
Latest advances fall into two broad categories: using ConL methods to optimize the representation space \cite{zhou-etal-2021-contrastive,zhou-etal-2022-knn,wang2023optimizing}, or employing generative modeling approaches like training a GPT model \cite{radford2019language} on IND data to achieve lower likelihoods for OOD text generation \cite{arora-etal-2021-types,wu-etal-2023-multi,ouyang-etal-2023-prefix}.

\noindent \textit{\textbf{T-3 New intent discovery, NID}}: 
NID is a classic and substantially studied task in NLP. Early studies focused on identifying new categories directly from unlabeled data, with most methods applying unsupervised clustering \cite{chatterjee-sengupta-2020-intent,an-etal-2023-dna}. 
The later approaches commonly incorporated some labeled data to support the discovery of unknown intents \cite{lin2020discovering,an2023new}. 
A typical workflow involves computing textual similarities between unlabeled and known data, assigning pseudo labels to unknown data based on the results, and then performing clustering. 
Recently, NID research has also emphasized the importance of RepL, with many studies incorporating pre-training and contrastive learning methods to optimize feature space arrangements \cite{zhang2021discovering,zhang-etal-2022-new,an2023generalized}.

\noindent \textit{\textbf{T-4 Continual learning}}: 
Continual learning generally denotes a re-training procedure to let the network learn new knowledge in stages, and finally reach a level to handle all tasks. For example, we may let a model learn the even five handwriting numbers and then the odd fives, finally let it recognize all ten numbers.
The biggest challenge of continual learning is catastrophic forgetting \cite{mccloskey1989catastrophic}, where the model's performance on earlier tasks deteriorates as it learns new tasks. Current studies have also been focusing on solving this issue, where the solutions mainly include: a) weight regularization \cite{paik2020overcoming,lin2022towards}, which imposes some constraints on the model's parameters to preserve values of model weights associated with knowledge from old data; b) knowledge distillation \cite{li2017learning,iscen2020memory}, which builds a smaller model to mimic the past model output as the hint for old tasks during handling new tasks; c) replay \cite{bonicelli2022effectiveness,gong-etal-2022-continual}, which involves storing a subset of old data and replaying it during training on new tasks to reinforce the old knowledge.

Although above methods have made much progress in their respective tasks, there still remains a gap in systematically integrating these tasks into a comprehensive QA system building, which is the aim of this paper.

\section{Task and workflow definition}
As mentioned in Section \ref{sec:intro}, we introduce four basic tasks to build a comprehensive and adaptive QA system (Figure \ref{fig:workflow}). Before presenting the specific solutions in the next section, we formally define these tasks and also simply introduce the workflow for better comprehension.

\subsection{Preliminary}
\noindent \textit{\textbf{Data}}: Suppose we initially have a set of known data $\mathcal{S}_{known}\text{=}\{\bm{x}_i, y_i | y_i $$\in$$ \mathcal{C}_{kn}\}$, where $\bm{x}_i$ is the \textit{i-th} user inquiry text, $y_i$ is the corresponding label, and $\mathcal{C}_{kn}$ is the label set of \textbf{known} intents. This data serves as the pre-training data, which is also referred as the In-Domain, IND examples, i.e., $\{\bm{x}_i, y_i|y_i $$\in $$\mathcal{C}_{kn}\}$$ \in $$\mathbb{D}_{IND}$.

We also have a set of label-unknown data $\mathcal{S}_{unlabeled}=\{\bm{x}_{j}, y_{j} | y_{j} $$\in$$ \{\mathcal{C}_{kn}$$+$$\mathcal{C}_{uk}\}\}$, which refers to the real, daily user inputs. Note that besides the known intents, there are also some data belonging to $\mathcal{C}_{uk}$, which is the \textbf{unknown} intent set. Those data are the so-called the Out-of-Domain, OOD examples, i.e., $\{\bm{x}_{j}, y_{j}|y_{j} $$\in$$ \mathcal{C}_{uk}\} $$\in$$ \mathbb{D}_{OOD}$.

\noindent \textit{\textbf{Encoder}}: We have a language encoder \textit{ENC} that encodes and transforms a discrete text $\bm{x}$ into an \textit{m}-dimension continuous representation $\bm{h}$: $\bm{x}$$\mapsto$$\bm{h}$$\in$$\mathbb{R}^{m}$.
Since there are many text similarity calculations during SCL training, we adopt a simple but effective ``mean-pooling'' strategy to obtain the sentence representation. Specifically, the average of all word embeddings in the sentence is served as the sentence embedding:

\begin{equation}
	\label{eq:sent-embd-cal}
	\small
	\boldsymbol{h}=\frac{1}{len(\boldsymbol{x})}\sum_{i=1}^{len(\boldsymbol{x})}ENC(w_i)
\end{equation}

\noindent where $len(\boldsymbol{x})$ is the length of input $\boldsymbol{x}$, $\{w_1,...,w_i,...,w_{len(\boldsymbol{x})}\}$ are the words in the sentence $\boldsymbol{x}$.

\subsection{T-1: User intent classification}
The target of this task is to find an accurate mapping:

\begin{equation}
	f_{cls}: ENC(\bm{x}_i) \mapsto \hat{y}_i \qquad \hat{y}_i \in \mathcal{C}_{kn}
\end{equation}

For traditional methods like CE-fine-tuning, this is achieved by further adding a fully-connected layer to the end of \textit{ENC} and training them together using CE loss. In the proposed method, \textit{ENC} is pre-trained using SCL (refer to Section \ref{ssec:scl} for details) and the parameters are fixed during T-1 and the following T-2 and T-3. The $f_{cls}$ is realized by a nearest neighbor algorithm (Section \ref{sssec:sol-t1}).

\subsection{T-2: Out-of-Domain detection}\label{ssec:problem-t2}
The target of this task is to determine whether an example $\bm{x}_{j}$ from $\mathcal{S}_{unlabeled}$ belongs to $\mathbb{D}_{IND}$ or $\mathbb{D}_{OOD}$. This is usually realized by comparing the detection score $sco$ of $\bm{x}_{j}$ with an empirically-set threshold $\lambda$, where $sco$ is given by a detection algorithm $f_{det}$ analyzed on the text embedding:

\begin{equation}
	sco_{j} = f_{det}(ENC(\bm{x}_{j}))
\end{equation}

\begin{equation}\label{eq:ind-ood-std}
	\bm{x}_{j} \in \left\{\begin{array}{l}\mathbb{D}_{OOD},\;if\;sco_{j}\geq \lambda\\  \mathbb{D}_{IND},\;if\;sco_{j}<\lambda\end{array}\right.
\end{equation}

In the actual workflow (as shown in Figure \ref{fig:workflow}), the detected $\bm{x}_{j'}^{ood}$ samples are saved for T-3 (new intent discovery), while the detected $\bm{x}_{j'}^{ind}$ examples will be classified normally in T-1 and also saved for the final task, T-4 (continual learning).

\subsection{T-3: New intent discovery}
Once the collected OOD texts $\bm{x}_{j'}^{ood}$ from T-2 reach a sufficient quantity, an algorithm $f_{nid}$ can be used to cluster and assign pseudo labels $\hat{y}_{j'}$ to these examples, as the ``new IND intent'' labels stored in set $\mathcal{C}_{kn'}$:

\begin{equation}
	f_{nid}: ENC(\bm{x}_{j'}^{ood}) \mapsto \hat{y}_{j'} \qquad \hat{y}_{j'} \in \mathcal{C}_{kn'}
\end{equation}

\subsection{T-4: Continual learning}\label{ssec:def-t4}
Finally, using both the known, classified IND data $\{\bm{x}_{j'}, \hat{y}_{j'} | \hat{y}_{j'} $$\in$$ \mathcal{C}_{kn}\}$ from T-1 and newly discovered ``IND'' data $\{\bm{x}_{j'}, \hat{y}_{j'} | \hat{y}_{j'} $$\in$$ \mathcal{C}_{kn'}\}$ from T-3, $ENC$ is re-trained to obtain the ability of classifying both old and new intent inquires.

\begin{equation}
	f_{cls\text{-}new}: ENC_{new}(\bm{x}_k) \mapsto \hat{y}_k \qquad \hat{y}_k \in \{\mathcal{C}_{kn}+\mathcal{C}_{kn'}\}
\end{equation}

%\noindent The above workflow is also shown in Figure \ref{fig:data-usage} for reference.

\section{Methodology}
In this section, we introduce the methodology of SCL as the RepL method in the pre-training stage, followed by the downstream methods to solve to each task.

\subsection{Stage 1: Supervised contrastive learning -based pre-training}\label{ssec:scl}
\subsubsection{Principle and optimizing effect of SCL}
As discussed in Section \ref{sec:related}, BERT+CE fine-tuning exhibits issues with local loose borders and a global anisotropic distribution in RepL, while SCL has the potential to solve both of these problems.

Specifically, as illustrated in Figure \ref{fig:framework}-b, for the local aspect, compared with CE fine-tuning, SCL applies a more straightforward yet effective target: pulling the representations of similar examples (as \textit{positive examples}) together and pushing dissimilar ones (as \textit{negative examples}) further apart. 
To tackle the global distribution problem, SCL further generates ``view'' augmentations as additional positive examples before the contrasting process (Figure \ref{fig:framework}-Pre-1), optimizing features in a boarder scale.
Together, these strategies provide a better supervision signal to arrange text representations to form compact and distinct clusters, enhancing both the alignment and uniformity of the feature space, and thus sharpening the distinctions between texts from various intents.

This well-organized embedding space provides a strong foundation that benefits all downstream tasks (as Figure \ref{fig:framework}-right part shows):
\begin{enumerate}[T-1:]
	\item The clear separations between each intent cluster facilitate easy and accurate classification.
	\item The compact and distinct IND clusters highlight the differences between them and OOD texts that do not belong to any known intents.
	\item Although not trained directly with OOD data, the two optimization directions of SCL still drive the aggregation of semantically similar features. This creates an environment conducive to intent discovery algorithms, making it easier to identify the pre-gathered data.
	\item The accurately discovered new intent data, pre-organized feature space, and SCL-based re-training, together ensure the continuous high-performance of the updated system.

\end{enumerate}

\subsubsection{Generation of text ``views''}
As discussed in the last subsection, the effective optimization of SCL is attributed to both the supervised positive-negative movement and the unsupervised ``view'' augmentation.
This augmentation also serves as a precaution against the lack of positive examples in the mini-batch, especially when the data quantity is small compared with category numbers.

For images, view generation can be achieved by simple transformations such as rotating and cropping. However, these techniques are not easily applicable to text ``view'' generation, as hasty word swaps or deletions can distort the original meaning.
In our previous study \cite{wang2023optimizing}, we found that encoding a sentence twice with different drop-out masks (Figure \ref{fig:framework}-Pre-1) is an effective way to generate high-quality, slightly different but main meaning preserved text embeddings. 
By using this generation method, SCL gave the best OOD detection results compared to no-view-generation \cite{zhou-etal-2021-contrastive} or more complex techniques like embedding-attack \cite{zeng-etal-2021-modeling} (refer to Tables 3-5 in \cite{wang2023optimizing} for detailed results). 
Considering these advantages, we continue to use this technique as the augmentation method in this paper.

\subsubsection{Formal learning procedure definition}\label{sssec:scl-process}
Here we give the formal definition of an SCL procedure:

\noindent \textit{\textbf{Step 1: Text ``View'' generation}}

Before the learning process, two text embedding views are generated for each example in the known dataset $\mathcal{S}_{known}\text{=}\{\bm{x}_i, y_i | y_i $$\in$$ \mathcal{C}_{kn}\}$. This is achieved by enabling the dropout option in network training settings and inputting the same sentence twice:
\begin{equation}
	\bm{h}_{2i'-1} = ENC_{drop1}(\bm{x}_i), \quad \bm{h}_{2i'} = ENC_{drop2}(\bm{x}_i)
\end{equation}

\noindent Here, since the two view embeddings are derived from the same text, we assign them the same label as $\bm{x}_i$, that is: we get two views $\{(\bm{h}_{2i'-1}, {y}_{2i'-1}),(\bm{h}_{2i'}, {y}_{2i'})\}$, where ${y}_{2i'-1}$ ${=}$${y}_{2i'}$${=}y_{i'}, i' $$\in$$ {1,2,...,N}$. $N$ is the total number of data.

\noindent \textit{\textbf{Step 2: Supervised contrastive learning}}

The SCL training loss is defined as follows:
\begin{equation}\label{eq-scl-loss}
	\small
	{\mathcal L}_{SCL}\text{=}\sum_{i'\text{=}1}^{2N}\frac{-1}{\left|Pos(i')\right|}\sum_{p\in Pos({i'})}\log\frac{exp(\boldsymbol {h}_{i'}\cdot \boldsymbol {h}_p/\tau)}{{\displaystyle\sum_{q\in Neg({i'})}}exp(\boldsymbol {h}_i'\cdot \boldsymbol {h}_q/\tau)}
\end{equation}

\noindent where $i'$$ \in $${1,2,...,2N}$ is the index of current sample (the \textit{anchor} text). $Pos({i'})$ is the index set of all \textit{positive} examples to the \textit{anchor} text in the mini-batch, i.e., $y_p\text{=}y_{i'}$, and $Neg({i'})$ is the index set of all \textit{negative} examples, i.e., $y_q$$\neq$$y_{i'}$. $\tau$ is the temperature factor to amplify the loss output.
To calculate text similarity directly using dot product, we apply $L2$ normalization to all representations $\boldsymbol {h}$ before training.

\subsection{Stage 2: Solution for each task after SCL}
With a well-organized embedding space achieved through SCL, now all four downstream tasks can be solved using simple methods with the least efforts.

\subsubsection{Solution for T-1 intent classification}\label{sssec:sol-t1}
For T-1, as shown in Figure \ref{fig:framework}-T-1, we apply a Mahalanobis distance (MDist) - nearest neighbor (NN)-based method to classify the text representations encoded by SCL-trained \textit{ENC}, which does not require any additional training. We choose MDist because, compared to Euclidean distance, MDist considers the distribution of each feature component, providing a more accurate measurement of text similarity.
Additionally, to further accelerate the classification speed but avoid the instability of NN, we propose calculating MDist between the current test sample $\boldsymbol x_{test}$ and only the centroids $\bar{\boldsymbol x}_c$ of each intent $c$. Finally, the label of one test example $y_{test}$ is determined as the label of the nearest centroid. The whole process can be described into the following three steps:

\noindent \textit{\textbf{a) Class centroid calculation}} 

The class centroid of category \textit{c} is the average data point of all text embeddings belonging to this intent:

\begin{equation}
	\bm{h}_{c,i} = ENC(\bm{x}_{c,i})
\end{equation}

\begin{equation}
	\small
	\bar{\boldsymbol h}_c= \frac1{num(c)}\sum_{i=1}^{num(c)} \boldsymbol h_{c,i}
\end{equation}

\noindent where $num(c)$ is data quantity of category \textit{c}, and $\boldsymbol h_{c,i}$ is the text representation of its \textit{i-th} example $\bm{x}_{c,i}$.

\noindent \textit{\textbf{b) MDist calculation}}

We calculate the covariance matrix $S$ over all data (eq. \ref{eq:cal-maha-cov}) and then calculate MDist (eq. \ref{eq-cal-maha}).

\begin{equation}\label{eq:cal-maha-cov}
	\small
	S=\frac1{\left|{\mathcal C}_{kn}\right|}\sum_{c\in {\mathcal C_{kn}}}(\boldsymbol H_c-\overline{\boldsymbol H}_c){(\boldsymbol H_c-\overline{\boldsymbol H}_c)}^T
\end{equation}
\begin{equation}\label{eq-cal-maha}
	\small
	\begin{aligned}
		MDist(\boldsymbol h_{test}, \bar{\boldsymbol h}_c)
		=\sqrt{{(\boldsymbol h_{test}-\bar{\boldsymbol h}_c)}^T S^{-1}(\boldsymbol h_{test}-\bar{\boldsymbol h}_c)}
	\end{aligned}
\end{equation}

\noindent where $\boldsymbol H_c \text{=} [\boldsymbol h_{c,1}, \boldsymbol h_{c,2}, ..., \boldsymbol h_{c,num(c)}]^T$ is the sample matrix of category $c$, $\overline{\boldsymbol H}_c\text{=} [1,1,..,1]^T\bar{\boldsymbol h}_c$.

\noindent \textit{\textbf{c) Label determination}}

Finally, the label is determined by the nearest centroid:
\begin{equation}\label{eq-label}
	\small
	y_{\text{test}} = \arg\min_{c \in \mathcal{C}_{kn}} MDist(\boldsymbol{h}_{\text{test}}, \bar{\boldsymbol{h}}_c)
\end{equation}

\subsubsection{Solution for T-2 out-of-domain detection}
The method to detect OOD examples is also very simple: we still calculate the embedding distance MDist between sample $\boldsymbol x_{test}$ and the nearest intent centroid as T-1, but the distance is used directly as the detection $sco$:

\begin{equation}\label{eq-sco}
	\small
	\begin{aligned}
		sco(\boldsymbol x_{test}) = {\underset{c\in {\mathcal C}_{kn}}{min}} \ MDist(\boldsymbol{h}_{\text{test}}, \bar{\boldsymbol{h}}_c)
	\end{aligned}
\end{equation}

This text representation distance itself can serve as the $sco$ and be an effective detection indicator, since a larger value means that even taking the shortest distance, current example is still far from, or say has less similarity to any known, IND category centroids. As shown in Figure \ref{fig:framework}-T-2, this suggests that the current $\boldsymbol x_{test}$ is more likely to be an unseen, i.e., OOD text. 
In practical usage, as introduced in task definition (Section \ref{ssec:problem-t2}), this $sco$ is further compared with a threshold $\lambda$, and texts with an exceeding distance will be judged as OOD examples.

\subsubsection{Solution for T-3 new intent discovery}\label{sssec:solution-t3}
To conduct new intent discovery, the first step is to obtain the unknown data to be clustered from T-2.
We achieve this by first finding a threshold $\lambda$ with a policy that when the model in T-2 gets a \textit{true positive rate, TPR=90\%} on validation data (i.e., 90\% OOD examples in validation set are correctly recognized), then we filter out OOD texts by comparing $sco$ with this $\lambda$.

Next, since SCL is able to provide a well-structured feature space, we do not conduct any further feature pre-processing or analysis but directly apply KMeans, one of the simplest unsupervised clustering method, as T-3's solution.
Taking the representations of detected OOD samples $\bm{x}_{j'}^{ood}$, KMeans clusters and assigns pseudo-labels $\hat{y}_{j'}$ to them. 
The detailed clustering procedure of KMeans in algorithm is provided in Appendix \ref{appsec:kmeans}. Finally, the pseudo-labeled new intent examples $\{\bm{x}_{j'}^{ood}, \hat{y}_{j'}\}$ will serve as the training data for the final task T-4.

\subsubsection{Solution for T-4 continual learning}
As mentioned in Section \ref{ssec:def-t4}, in the final task, the initial model $ENC$ is re-trained with both detected IND examples $\{\bm{x}^{ind}_{j'}, \hat{y}_{j'} | \hat{y}_{j'}$$ \in $$\mathcal{C}_{kn}\}$ (labeled by T-1) and detected OOD examples $\{\bm{x}^{ood}_{j'}, \hat{y}_{j'} | \hat{y}_{j'} $$\in$$ \mathcal{C}_{kn'}\}$ (pseudo-labeled in T-3) to obtain the ability to classify new intents. The re-training method employed here remains SCL (as in Stage-1, Section \ref{sssec:scl-process}) but with an expanded label set $\{\mathcal{C}_{kn}$$+$$\mathcal{C}_{kn'}\}$ and without further modifications.

This training method can be regarded as a straightforward ``replay'' method to mitigate forgetting of the old intents. We choose this strategy because the goal of continual learning in our QA system is to enable the model to classify all categories (both old and new intents) simultaneously, rather than in separate stages like other studies. Additionally, since the old intent data are readily available and can be stored without significant cost, as a result, the replay strategy is simple, easy-to-deploy and well-suited to our objectives.
Experiment results (Section \ref{ssec:result-t4}) also demonstrate that this replay policy effectively prevents forgetting, with only minimal performance degradation observed in classifying old data.

\section{Experiment}
\subsection{Datasets, division, and evaluation order}
To clearly present the evaluation process for the four tasks, we begin by introducing the original dataset, then specify the data divisions applied for the experiment, and finally outline the evaluation workflow based on these divisions.

\subsubsection{Dataset}
We use three popular public intent classification datasets for the experiment.

\noindent \textit{\textbf{CLINC}} \cite{larson-etal-2019-evaluation}: 
This is a widely-used dataset in intent classification and OOD detection tasks. It provides 150 intents of texts and one OOD intent data (that is, on OOD detection task, the former 150 intents can be treated as the IND intents). Because its OOD part does not contain specific labels that are necessary for T-3 and T-4 evaluation, we only use its 150-intent part data this time.

\noindent \textit{\textbf{BANKING}} \cite{casanueva-etal-2020-efficient}: This is an intent classification dataset concerning online banking services, which contains around 13,000 records in 77 intents. 

\noindent \textit{\textbf{SNIPS}} \cite{Coucke2018}: This is another intent classification dataset extracted from a voice assistant, consisting of around 14,000 records in 7 intents.

The original data composition of above datasets is shown in Table \ref{tab:ori-data}. Some examples are also provided in Table \ref{tab:data-eg}.
Since all data are well-formed, which keep a relatively good data balance and do not contain personal information, we did not take further pre-processing steps but used them as they are.  

\begin{table}[t]
	\centering
	\caption{Original data composition of each dataset}
	\begin{tabular}{lcccc}
		\toprule
		Dataset & num. of cate. & Train & Val & Test \\
		\midrule
		CLINC & 150   & 15,000 & 3,000  & 4,500 \\
		BANKING & 77    & 9,003  & 1,000  & 3,080 \\
		SNIPS & 7     & 13,000 & 700   & 700 \\
		\bottomrule
	\end{tabular}%
	\label{tab:ori-data}%
\end{table}%

\begin{table}[t]
	\centering
	\caption{Examples of each dataset}
	\resizebox{0.98\linewidth}{!}{
		\begin{tabular}{m{7em}<{\centering}m{20em}<{\centering}}
			\toprule
			Intent name & Example \\
			\midrule
			\multicolumn{1}{c}{\textbf{CLINC}} &  \\
			translate & what is the equivalent of, 'life is good' in french \\
			distance & how many miles will it take to get to my destination \\
			cancel\_reservation & please cancel the table for two at burger king \\
			\midrule
			\multicolumn{1}{c}{\textbf{BANKING}} &  \\
			verify\_source\_ of\_funds & Where can I check out the source of available funds in my account? \\
			Refund\_not\_ showing\_up & I contacted the seller for a direct refund last week but he still hasn't give me my money. What should I do? \\
			cash\_withdrawal\_ not\_recognised & Hello, Please look into this matter urgently. As i have lost my wallet and i can see one withdrawal, now i do not want to loose more money. \\
			\midrule
			\multicolumn{1}{c}{\textbf{SNIPS}} &  \\
			BookRestaurant & book a restaurant at sixteen o clock in sc \\
			SearchScreeningEvent & find me showtimes for animated movies in the neighbourhood \\
			AddToPlaylist & put this song in my funk playlist please \\
			\bottomrule
	\end{tabular}}%
	\label{tab:data-eg}%
\end{table}%

\subsubsection{Data division for the experiment}
We apply two key divisions to adapt the data for the experimental requirements. 
The first is the IND/OOD intent split, where approximately 75\% number of the original intents are randomly selected as the known IND intents, with the remaining 25\% treated as the unknown OOD intents. 
Specifically, in \textit{CLINC, BANKING} and \textit{SNIPS} datasets, 112, 57, and 5 intents were randomly selected as the known IND intents, respectively, while the remaining 38, 20, and 2 intents were designated as the unknown OOD intents.

Additionally, since T-4 requires the recognition results from T-2 and T-3 for the final re-training (refer to Section \ref{ssec:def-t4} for this process and Figure \ref{fig:data-usage} for workflow), the original single test set is insufficient to evaluate all four tasks, especially as it may cause data leakage for T-4. To address this, we further split around 20\% data from the original training set to create a secondary test set, Test II.
The new composition of the re-divided data is presented in Table \ref{tab:re-data}.

\begin{table}[bt]
	\centering
	\caption{Re-organized data composition}
	\begin{tabular}{lccccc}
		\toprule
		Dataset &  IND/OOD cate. & Train & Val & Test I & Test II\\
		\midrule
		CLINC & 112/38   & 12,000 & 3,000  & 4,500 & 3,000\\
		BANKING & 57/20    & 7,463  & 1,000  & 3,080 & 1,540\\
		SNIPS & 5/2     & 12,300 & 700   & 700 & 700\\
		\bottomrule
	\end{tabular}%
	\label{tab:re-data}%
\end{table}%

\subsubsection{Evaluation order and data usage}\label{sssec:data-use}
Here we introduce the training and evaluation processes using above divided data. A simple workflow diagram is shown in Figure \ref{fig:data-usage}, while the complete procedure in pseudo-code form can also be found in Appendix \ref{appsec:data-usage}.

\noindent \textbf{Training stage}: As shown in Figure \ref{fig:data-usage}, in the \textbf{pre-training} stage, the model is trained with an SCL target using only the IND part of training data (Train-IND). During training, Val - IND/OOD data are used to determine the stopping point, specifically when the OOD detection AUROC (refer to Section \ref{ssec:metric} for definition) on Val set reached its peak.

\noindent \textbf{Evaluation stage}: After pre-training, the intent classification performance (\textbf{T-1}) of the initial model is first tested on Test I-IND data. The OOD detection ability (\textbf{T-2}) is tested on Test I-IND/OOD data. After that, the filtered OOD texts from T-2 (refer to Section \ref{sssec:solution-t3} for filtering strategy) are used for new intent discovery (\textbf{T-3}) and the accuracy is then evaluated. 
Finally, using the pseudo-labeled old IND data (labeled by initial model) and new ``IND'' data (labeled in T-3), the initial model is re-trained (\textbf{T-4}), and the classification performance is tested on Test II set.

\begin{figure}[tb]
	\centering
	\includegraphics[width=0.9\linewidth]{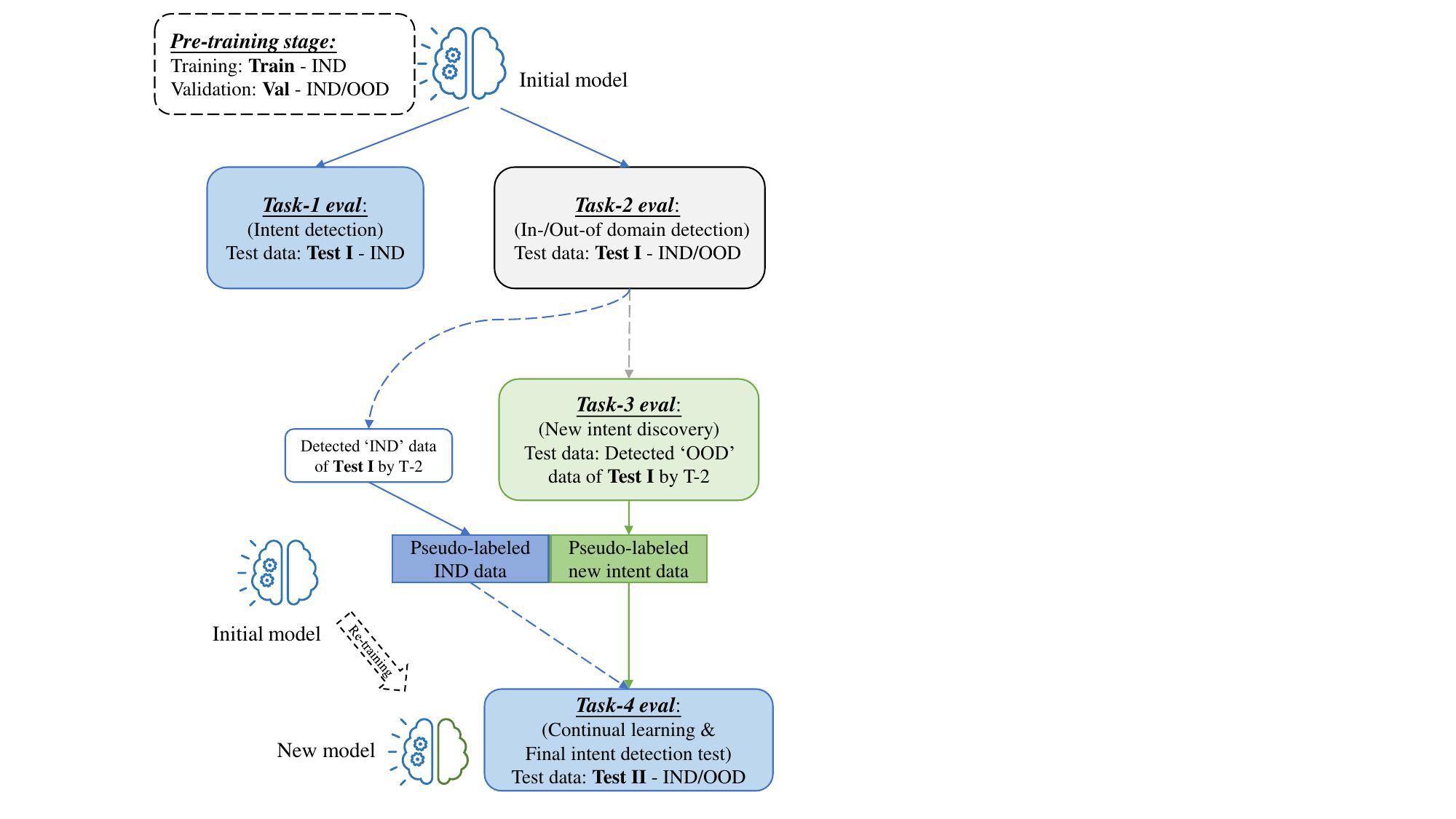}
	\caption{Diagram of the training and evaluation workflow}
	\label{fig:data-usage}
\end{figure}

\subsection{Metrics}\label{ssec:metric}
We use the following metrics for model evaluation in each task. Basic concepts are provided here, with detailed calculation processes in Appendix \ref{appsec:metric}.

\subsubsection{For Task 1 and 4: F1 values}
Since T-1 and T-4 both belong to classification problems, we use \textit{\textbf{Micro}} and \textit{\textbf{Macro F1}} scores for the evaluation.

\subsubsection{For Task 2: AUROC, AUPR, FPR90}

As implied in Section \ref{ssec:problem-t2}, the accuracy of IND/OOD detection is influenced by the comparison between $sco$ and different $\lambda$. To ensure a fair and standard evaluation, we use three threshold-independent classification metrics.

\noindent \textit{\textbf{AUROC/AUPR}}: AUROC, or \textit{Area Under the Receiver Operating Characteristic curve}, measures the area under the curve of true positive rate (TPR) against false positive rate (FPR) at varying $\lambda$ values. AUPR stands for \textit{Area Under Precision-Recall curve}. Higher values in either metric indicate better OOD detection performance.

\noindent \textit{\textbf{FPR90}}:
A practical metric which shows the FPR value when TPR reaches 90\%. Lower values indicate better performance.

\subsubsection{For Task 3: NMI, ARI, ACC}
For T-3, we use three common metrics in clustering and intent discovery: Normalized Mutual Information (\textit{\textbf{NMI}}), Adjusted Rand Index (\textit{\textbf{ARI}}) and Clustering Accuracy (\textit{\textbf{ACC}}). 
Simply speaking, NMI and ARI measure the similarity between two clusters, while ACC evaluates the clustering accuracy after finding the best true - pseudo-clustering label mapping.
Higher values in these metrics indicate a better performance.

\subsection{Baselines}
We apply the following latest and competitive methods as baselines for evaluation.

\noindent \textit{\textbf{Fine-tuning, FT}} (For all four tasks): Despite being the most common training method, CE-based fine-tuning remains one of the most powerful methods actively used among all tasks \cite{wang2021legal,podolskiy2021revisiting,lin2020discovering}. Thus this time we applied it as a baseline of RepL, and tested the effects throughout all four tasks. After fine-tuning, the downstream detection and clustering methods are all the same as SCL.

\noindent \textit{\textbf{PTO-Sup}} \cite{ouyang-etal-2023-prefix} (For T-2): This work proposed a Prefix Tuning-based OOD detection method, training a generative GPT-2 model on IND data and then using the likelihood as $sco$. It assumes that the unseen OOD text should let model output a lower likelihood. This method originally supports unsupervised learning, but for a fair comparison, we applied its supervised mode with IND text labels available.

\noindent \textit{\textbf{Layer-LOF}} \cite{zhou-etal-2023-two} (For T-2): This paper addressed that pre-trained language models like BERT may over-rationalize the input. That is, even it is an OOD input, the last-layer embedding may tend to be more like IND. Based on this observation, it proposed to consider a decision of multiple layers after fine-tuning, and finally used local outlier factor (LOF) with a voting mechanism for the final decision. In this paper, we simply averaged its best-four layer results as the final result.

\noindent \textit{\textbf{MTP-KNNCL}} \cite{zhang-etal-2022-new} (For T-3): The state-of-the-art NID method that first pre-trains the model with a Masked Language Modeling (MLM) target to obtain prior semantic knowledge, then performs KNN contrastive learning for better cluster aggregation. The downstream clustering method is KMeans.

\noindent \textit{\textbf{DPN}} \cite{an2023generalized} (For T-3): This Decoupled Prototypical Network first initializes some prototypes for known and unknown intents and aligns them as a bipartite matching problem. Unaligned prototypes represent new intents, and all embeddings are further optimized using semantic-aware prototypical learning. The downstream method is also KMeans.

\begin{table*}[t]
	\centering
	\caption{T-1: intent classification result on Test I - IND set}
	\resizebox{0.98\linewidth}{!}{
		\begin{tabular}{cccccccccc}
			\toprule
			Dataset & \multicolumn{3}{c}{\textbf{CLINC}} & \multicolumn{3}{c}{\textbf{BANKING}} & \multicolumn{3}{c}{\textbf{SNIPS}} \\
			\midrule
			Method/Metric & Micro F1$\uparrow$ & Macro F1$\uparrow$ & Time/Epoch & Micro F1$\uparrow$ & Macro F1$\uparrow$ & Time/Epoch & Micro F1$\uparrow$ & Macro F1$\uparrow$ & Time/Epoch \\
			\midrule
			\textit{FT-FullyConnect} & \textbf{97.23} & 97.22 & 4min10s/18 & 87.32 & 85.69 & 3min/14 & \textbf{98.83} & 98.74 & 3min15s/15\\
			\textit{SCL-NN (Ours)} & \textbf{97.23} & \textbf{97.23} & \textbf{1min10s}/7 & \textbf{94.17} & \textbf{94.16} & \textbf{2min30s}/10 & \textbf{98.83} & \textbf{98.75} & \textbf{2min40s}/12 \\
			\bottomrule
	\end{tabular}}%
	\label{tab:result-t1}%
\end{table*}%

\begin{table*}[t]
	\centering
	\caption{T-2: IND/OOD detection result on Test I - IND/OOD set, the best result is in \textbf{bold}, the runner-up result is \underline{underlined}}
	\resizebox{0.98\linewidth}{!}{
		\begin{tabular}{ccccccccccccc}
			\toprule
			Dataset & \multicolumn{4}{c}{\textbf{CLINC}} & \multicolumn{4}{c}{\textbf{BANKING}} & \multicolumn{4}{c}{\textbf{SNIPS}} \\
			\midrule
			Method/Metric & AUROC$\uparrow$ & AUPR$\uparrow$  & FPR90$\downarrow$ & Time/Epoch  & AUROC$\uparrow$ & AUPR$\uparrow$  & FPR90$\downarrow$ & Time/Epoch  & AUROC$\uparrow$ & AUPR$\uparrow$  & FPR90$\downarrow$ & Time/Epoch \\
			\midrule
			\textit{PTO-Sup} & 81.67 & 55.01 & 44.3  & 40min/8 & 79.36 & 56.44 & 55.0  & 72min/16 & \textbf{97.53} & \underline{93.87} & \underline{6.4}   & 3min20s/13 \\
			\textit{Layer-LOF} & \underline{95.67} & \underline{86.71} & 10.7  & 78min/19 & \underline{92.89} & \underline{81.67} & 20.3  & 47min/12 & 93.47 & 87.34 & 15.6  & 28min/8 \\
			\textit{FT-NDist} & 95.64 & 85.96 & \underline{9.7}   & \underline{4min10s}/18 & 92.73 & 80.12 & \underline{18.9}  & \underline{3min}/14  & 96.56 & 90.43 & 9.2   & \underline{3min15s}/15 \\
			\textit{SCL-NDist (Ours)} & \textbf{96.15} & \textbf{88.41} & \textbf{8.5} & \textbf{1min10s}/7 & \textbf{93.44} & \textbf{82.06} & \textbf{15.7} & \textbf{2min30s}/10 & \underline{97.41} & \textbf{94.36} & \textbf{4.3} & \textbf{2min40s}/12 \\
			\midrule
			\multicolumn{13}{l}{Note: All results are calculated under `OOD example as class 1 (positive) and IND example as class 0 (negative)' condition to highlight the model's detection ability. }\\
	\end{tabular}}%
	\label{tab:result-t2}%
\end{table*}%

\begin{table*}[t]
	\centering
	\caption{T-3: New intent discovery result on Test I - OOD set, the best result is in \textbf{bold}, the runner-up result is \underline{underlined}}
	\resizebox{0.98\linewidth}{!}{
		\begin{tabular}{ccccccccccccc}
			\toprule
			Dataset & \multicolumn{4}{c}{\textbf{CLINC}} & \multicolumn{4}{c}{\textbf{BANKING}} & \multicolumn{4}{c}{\textbf{SNIPS}} \\
			\midrule
			Method/Metric & NMI$\uparrow$   & ARI$\uparrow$   & ACC$\uparrow$   & Time/Epoch  & NMI$\uparrow$   & ARI$\uparrow$   & ACC$\uparrow$   & Time/Epoch  & NMI$\uparrow$   & ARI$\uparrow$   & ACC$\uparrow$   & Time/Epoch \\
			\midrule
			\textit{MTP-KNNCL} & \underline{92.36} & \underline{79.70}  & \underline{83.60}  & 6min10s/20 & \underline{83.33} & \underline{69.75} & \underline{80.75}  & 6min15s/20 & \textbf{78.04} & \textbf{85.50} & \textbf{96.26} & 6min40s/20 \\
			\textit{DPN} & 89.93 & 75.19 & 82.89 & 6min30s/20 & 82.51 & 69.04 & 76.50  & 6min/20  & 68.85 & \underline{77.72} & \underline{94.12} & 6min30s/20 \\
			\textit{FT-KMeans} & 86.81 & 67.71 & 76.27 & \underline{4min10s}/8 & 81.88 & 64.88 & 73.86 & \underline{3min}/14  & 66.16 & 70.29 & 91.98 & \underline{3min15s}/15 \\
			\textit{SCL-KMeans (Ours)} & \textbf{93.08} & \textbf{81.14} & \textbf{85.67} & \textbf{1min10s}/7 & \textbf{84.44} & \textbf{70.64} & \textbf{81.16} & \textbf{2min30s}/10 & \underline{70.94} & 75.82 & 93.58 & \textbf{2min40s}/12 \\
			\bottomrule
	\end{tabular}}%
	\label{tab:result-t3}%
\end{table*}%

\subsection{Other implementation details}
We adopted \textit{BERT-base} \cite{devlin2018bert} and \textit{MPNet-base} \cite{song2020mpnet} as the base encoders for both our methods and baselines (expect for \textit{PTO-Sup} that used GPT-2 small \cite{radford2019language}). These encoders were chosen because BERT is the most commonly used pre-trained encoder, while MPNet, a BERT-like variation, enhances context learning ability for better sentence encoding \cite{reimers-gurevych-2019-sentence}.
For SCL setting, we set dropout rate $p\textit{=}0.1$, $n_\text{views}\textit{=}2$ for ``view'' generation, and $\tau\textit{=}0.1$ for calculating the loss. 
During the pre-training stage, we used the OOD detection AUROC on Val set to evaluate the training process for our \textit{SCL} and \textit{FT} methods, saving only the model with the highest $AUROC_{val}$. The maximum training epoch was set to 20 for both our and all baseline methods.
All experiments were conducted three times, and the results were averaged to obtain the final result. All experiments were running on the same machine with an Ubuntu 18.04 operating system, an Nvidia A6000 GPU, an Intel i9-10980X CPU, and 128GB RAM.

\section{Results and Analysis}\label{sec:result}
\subsection{Result on task 1: user intent classification}
The intent classification results on Test I-IND data are shown in Table \ref{tab:result-t1}. We also provide the training time and epoch as an efficiency reference. 
It can be found that both proposed \textit{SCL} and traditional \textit{FT} method achieved very high F1 scores on CLINC and SNIPS datasets. 
However, things changed on BANKING, where \textit{FT} method saw a considerable accuracy drop in both metrics, while our method kept a relatively good performance. 
We analyze that this is because, compared with the other two datasets, the user inquires in BANKING have a longer input length in average, and the inquiry expressions also become more diverse (that is, more detailed and have many patterns, as the examples shown in Table \ref{tab:data-eg}), which greatly increased classification difficulty. 
Despite this, the proposed \textit{SCL-NN} method was still able to give a more inter-class separated and intra-class compact embedding space (a visual prove will be given in Section \ref{ssec:result-tsne}), guaranteeing a high accuracy even with longer and noisier inputs.

\subsection{Result on task 2: out-of-domain detection}\label{ssec:result-t2}
The results evaluating the detection ability of unknown OOD inputs are shown in Table \ref{tab:result-t2}. 
Firstly, comparing \textit{FT} and \textit{SCL}, it can be observed that while \textit{FT} provided very solid results, the proposed method performed even better and achieved nearly all the best results. 
As for the other baselines, \textit{Layer-LOF} that applied a layer-voting strategy obtained only slightly better results than \textit{FT} on the first two datasets. Nevertheless, the cost is a much longer training time by the extensive computations required to measure the information redundancy between layers.
The random layer selection strategy also increased uncertainty, making it less practical. 
On the other hand, the generative model-based method \textit{PTO-Sup} worked badly except on SNIPS, which even obtained the highest AUROC on it. We speculate the reason of this abnormal situation as compared with the other two datasets, the IND/OOD examples in SNIPS exhibited more ``background-shift'' than usual ``semantic-shift'', which can be exactly the strong point of generation-based methods.

Talking more specifically, for instance, there are two intents belonging to IND and OOD intents shared the topic of ``watching a movie'' in SNIPS, although one was looking for movie information and another involved planning a cinema visit, which is a typical ``background-shift''. (On the contrary, ``watching a movie'' and ``read a book'' intent pair is totally different in topic and thus is a ``semantic-shift''.)
This difference is easier for generative and textual perplexity-based methods like \textit{PTO-Sup} to capture and distinguish, while it can be a weak point of CE fine-tune-based methods \cite{arora-etal-2021-types}. This is also evident from the performance drop of the other two fine-tuning-based baselines on SNIPS. 
However, it is clear that \textit{SCL} still held strong on not only the ``semantic-shift'', but also the ``background-shift'' situation, showing its superiority on the OOD intent detection task.

\begin{table*}[tbp]
	\centering
	\caption{T-4: Continual learning result on Test II set}
	\begin{tabular}{ccccccc}
		\toprule
		Dataset & \multicolumn{2}{c}{\textbf{CLINC}} & \multicolumn{2}{c}{\textbf{BANKING}} & \multicolumn{2}{c}{\textbf{SNIPS}} \\
		\midrule
		Method/Metric & Micro F1$\uparrow$ & Macro F1$\uparrow$ & Micro F1$\uparrow$ & Macro F1$\uparrow$ & Micro F1$\uparrow$ & Macro F1$\uparrow$ \\
		\midrule
		OVERALL &       &       &       &       &       &  \\
		\textit{FT} & 92.65 & 93.37 & 79.55 & 81.69 & 94.79 & 94.86 \\
		\textit{SCL} & \textbf{97.70 } & \textbf{97.69} & \textbf{91.75} & \textbf{91.79} & \textbf{99.14} & \textbf{99.14} \\
		\midrule
		ON IND &       &       &       &       &       &  \\
		\textit{FT} & 97.53 & 97.48 & 86.29 & 84.33 & 98.80  & 98.80 \\
		\textit{SCL} & \textbf{98.14} & \textbf{98.10 } & \textbf{93.53} & \textbf{93.55} & \textbf{99.30} & \textbf{99.29} \\
		Ref.-Initial FT model & 98.30  & 98.31 & 87.11 & 85.93 & 99.40  & 99.40 \\
		Ref.-Initial SCL model & 98.39 & 98.40  & 93.60  & 93.63 & 99.40  & 99.40 \\
		\midrule
		ON OOD &       &       &       &       &       &  \\
		\textit{FT} & 89.53 & 85.65 & 82.14 & 78.04 & 92.18 & 91.52 \\
		\textit{SCL} & \textbf{98.48} & \textbf{98.46} & \textbf{92.51} & \textbf{92.30} & \textbf{99.75} & \textbf{99.75} \\
		\bottomrule
	\end{tabular}%
	\label{tab:result-t4}%
\end{table*}%

\subsection{Result on task 3: new intent discovery}\label{ssec:result-t3}
Table \ref{tab:result-t3} shows the results for new intent discovery. 
Comparing \textit{SCL} and traditional \textit{FT} method, the advantage of \textit{SCL} concerning feature optimization capability is proven again by the surpassing clustering scores on all datasets.
As for the other baselines, \textit{DPN} also achieved better scores than \textit{FT}, but the improvement was not significant. \textit{MTP-KNNCL} showed very strong clustering performance, achieving second best results to our method on the first two datasets and the best results on SNIPS that other methods performed much worse. 

We consider the possible reason as the existence of ``background-shift'' in SNIPS discussed in T-2 result, where the supervised training style of the other three methods might exacerbate the confusion between those intents. On the other hand, the two techniques, unsupervised MLM target and semi-supervised KNN-ConL, adopted by \textit{MTP-KNNCL} mitigated the risk of ``over-training'' in supervised settings, and thus giving a better result. However, the cost is a slightly longer training time.

\subsection{Result on task 4: continual learning}\label{ssec:result-t4}
Finally, we present the intent classification results after the incremental learning in Table \ref{tab:result-t4}. From an overall view (results on all classes), \textit{SCL} has continued to achieve the best results, and even widened the accuracy gap with \textit{FT} on all datasets from 0 - 8.5\% (on T-1 initial classification) to 4.3\% - 10.1\%, which we consider as a result of the excellent clustering performance on T-3 and the feature optimization ability of \textit{SCL}.
The concerning forgetting phenomenon also did not appear, as only a slight accuracy decrease on the old IND intent classification was observed after the continual learning. This proves that the adopted ``replay'' strategy using both pseudo-labeled IND and OOD texts is effective.

In addition to the continuous high accuracy, another advantage of \textit{SCL}-based re-training is also worth noting. 
When conducting \textit{FT}-based re-training, the last fully-connected layer had to be replaced and reinitialized every time to accommodate the new intent number, and it required at least 4-5 epochs to obtain desirable results in our experiments. 
On the other hand, since \textit{SCL} focuses on the feature optimization and the label judgment is based on the nearest neighbor algorithm, it can directly use the new data with an extended label set for model re-training in a shorter time.  This easy-to-use expandability can save a considerable amount of resources and is meaningful in real-world deployments.

\begin{figure*}[t]
	\centering 
	\subfigure[SCL (all test i data)]{
		\label{level.sub.1}
		\includegraphics[width=0.27\linewidth]{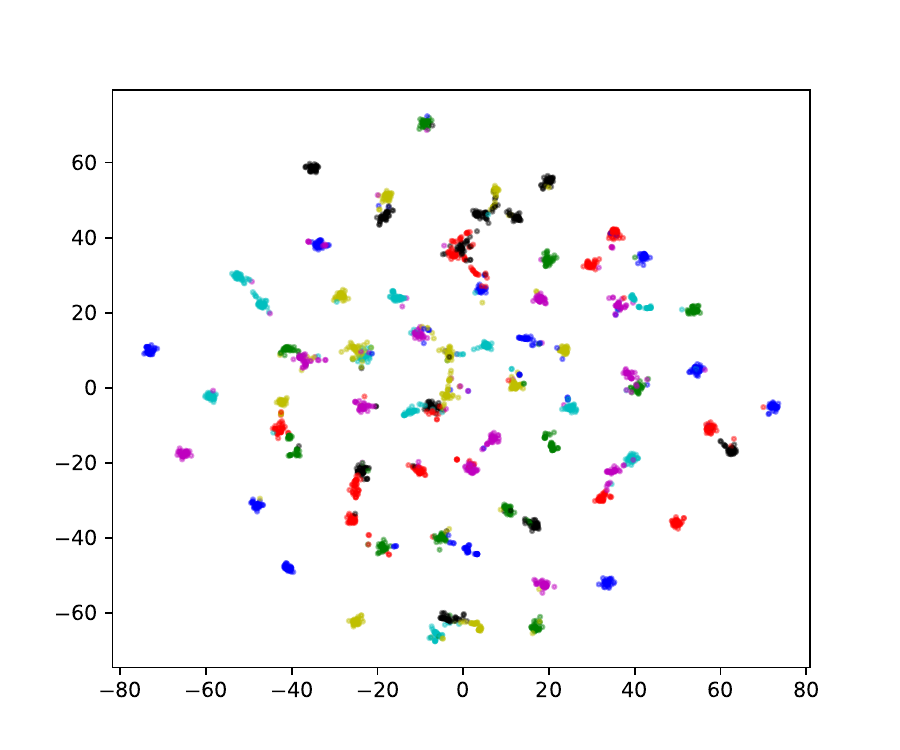}}
	\quad 
	\subfigure[FT (all test i data)]{
		\label{level.sub.2}
		\includegraphics[width=0.27\linewidth]{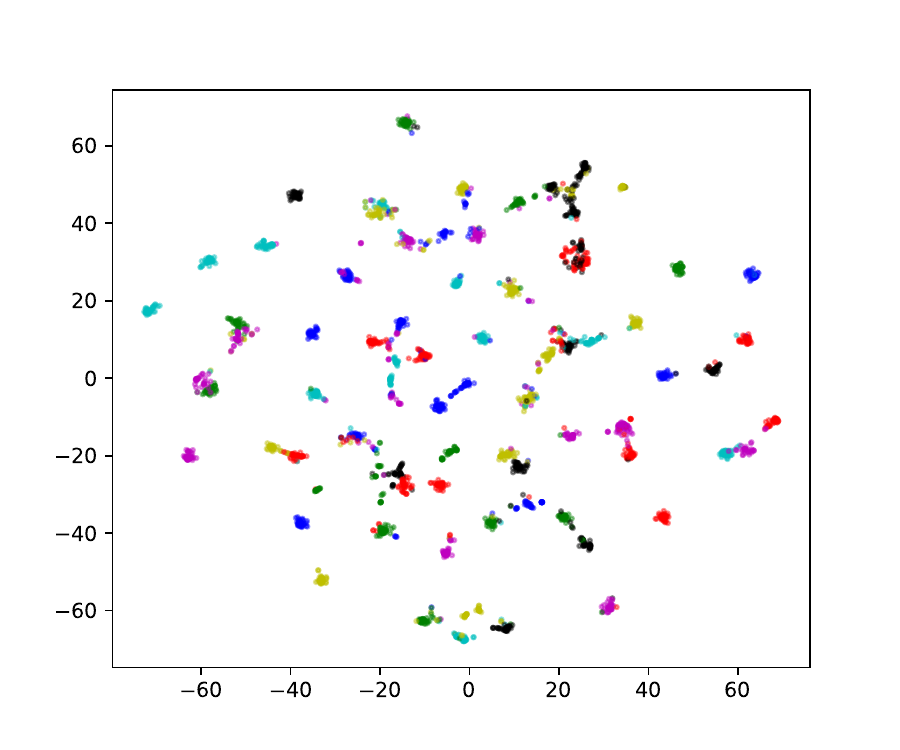}}
	\quad 
	\subfigure[MTP-KNNCL (all test i data)]{
		\label{level.sub.3}
		\includegraphics[width=0.27\linewidth]{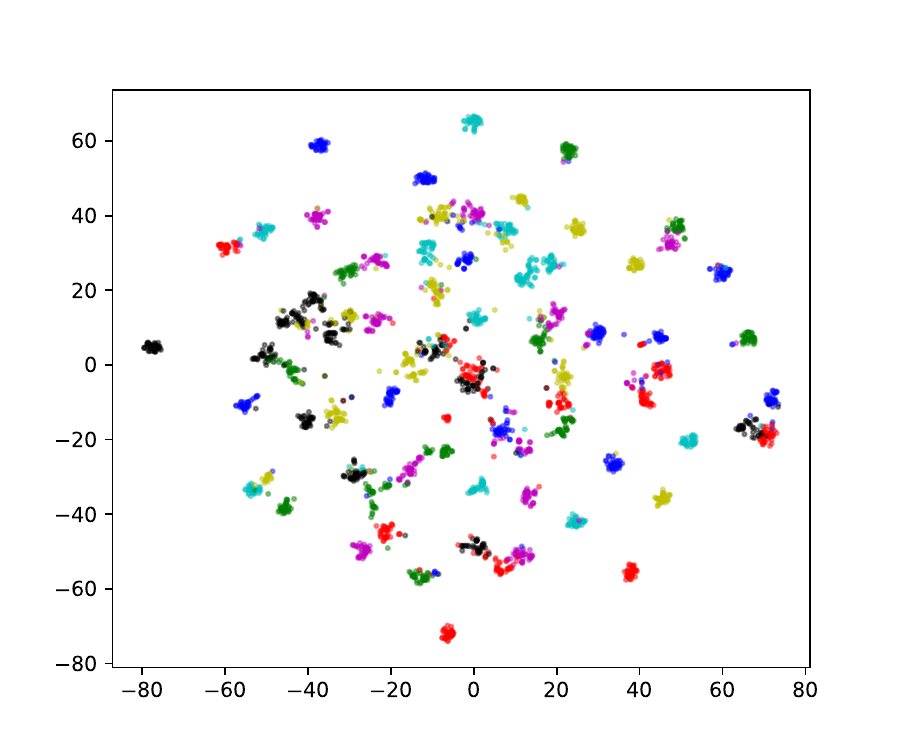}}

	\subfigure[SCL (test i - ood data)]{
		\label{level.sub.4}
		\includegraphics[width=0.27\linewidth]{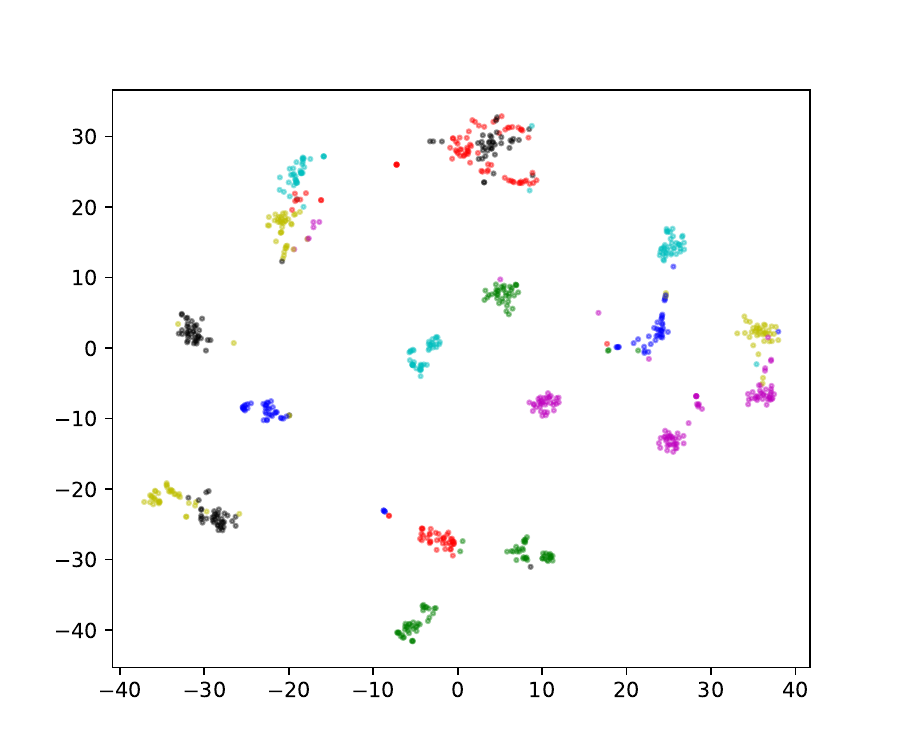}}
	\quad
	\subfigure[FT (test i - ood data)]{
		\label{level.sub.5}
		\includegraphics[width=0.27\linewidth]{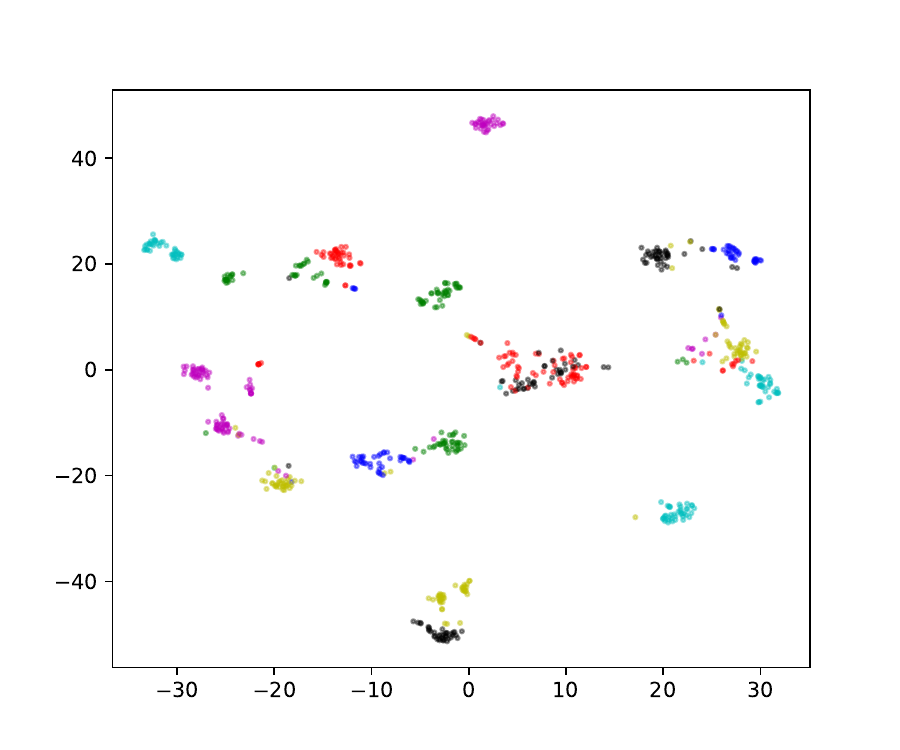}}
	\quad
	\subfigure[MTP-KNNCL (test i - ood data)]{
		\label{level.sub.6}
		\includegraphics[width=0.27\linewidth]{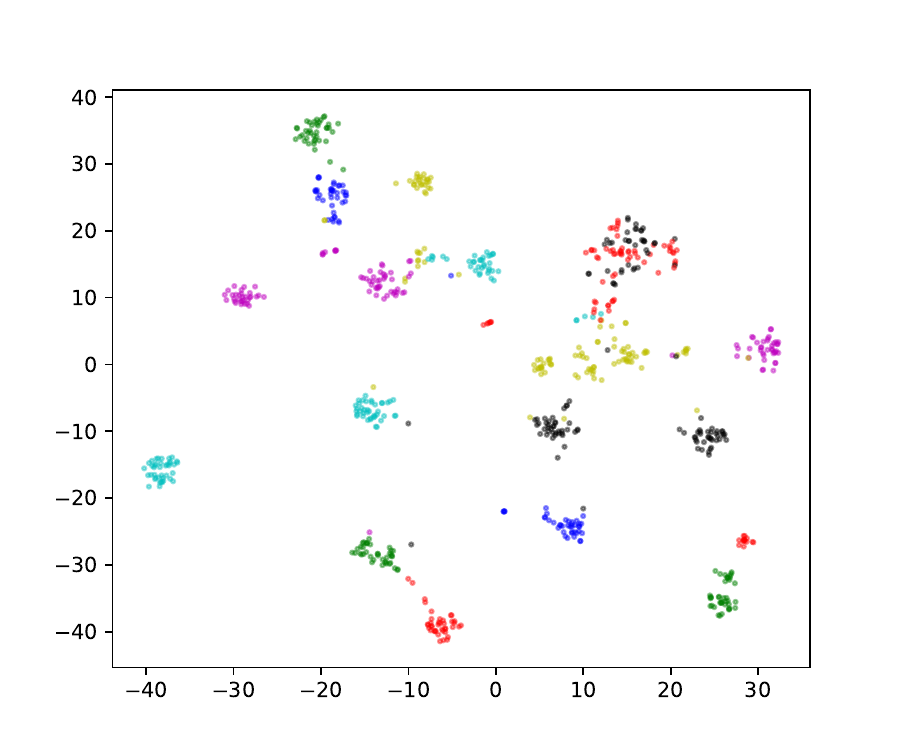}}
	\caption{T-SNE text feature space visualization result on BANKING test I - all data (top row) and only OOD data (bottom row)}
	\label{fig:t-sne}
\end{figure*}

\section{Additional studies and discussions}

\subsection{Embedding space visualization}\label{ssec:result-tsne}
To verify the representation optimization effect, we conducted a t-SNE embedding space visualization on BANKING Test I set, which has a moderate number of examples and categories to view results easily. The results are shown in Figure \ref{fig:t-sne}. The top row displays the embedding arrangement over all intents, while the bottom row focuses only on the newly discovered (i.e., previous OOD intents) data.

From the visualization results on overall data distribution (Figure \ref{fig:t-sne} a-c), it can be observed that both \textit{SCL} and \textit{FT} performed reasonably well, but the embedding space generated by \textit{FT} showed more mixed embeddings from different intents. This proves that our proposed method has a better cluster aggregation effect over traditional \textit{FT}. 
\textit{MTP-KNNCL} produced relatively larger clusters in the space with some clusters very close to each other, but the overall distribution situation is slightly better than \textit{FT}.
We consider this may be due to the optimization style of KNN-ConL part used in \textit{MTP-KNNCL}, which continuously pulled nearby similar embeddings together, resulting in this side effect on cluster aggregation. 

From the visualization results of newly discovered intents (Figure \ref{fig:t-sne} d-f), the advantage of proposed method is more obvious. 
Without accessing OOD texts, \textit{SCL} was still able to separate most OOD classes, with only two class clusters apparently mixed. \textit{MTP-KNNCL} also worked relatively well, but the clusters were gathered closer, which showed an inferior labeling dispersion effect to our method. Finally, \textit{FT} produced a much flatter embedding distribution with more severe confusion between intents. These visualization results correspond exactly to the results in Task 3.

To conclude, the visualization result proved again that despite using less than half the time to the state-of-the-art method \textit{MTP-KNNCL} and requiring fewer tuning steps, our method still produced a splendid underlying embedding space for clustering and other tasks, which is valuable in practical applications.

\subsection{Computational efficiency analysis}
Computational efficiency refers to the time or resources required for model training. In this analysis, we use the total computation amount required for a complete training procedure as an indicator of computational efficiency, which can be represented as:

\begin{equation}\label{eq:eff}
	Eff.\propto Amount\;=\;Epoch\ast(O(model)+O(loss))
\end{equation}

\noindent where \textit{O(model)} denotes the complexity of model's forward and back propagation, and \textit{O(loss)} represents the computation complexity of the loss function.

Since all methods in the experiment used the same base encoder, the \textit{O(model)} part of them can be considered to be approximately equal\footnote{Although GPT-2-small and BERT-base differ, their parameter sizes are close, so we consider their \textit{O(model)} to be nearly equal.}. In addition, given that \textit{O(model)} is usually significantly larger than \textit{O(loss)} (The analysis of \textit{O(loss)} of each method and \textit{O(model)} can be found in Appendix \ref{app:eff}), eq. \ref{eq:eff} can be simplified to:

\begin{equation}\label{eq:eff-simple}
	\begin{split}
		& Eff.\propto Amount \propto Epoch\ast O(model) \\
		& \rightarrow Eff.\propto Amount \propto Epoch
	\end{split}
\end{equation}

This simplification indicates that the computational efficiency, represented by computation amount, is primarily related to the number of training epochs. Reviewing the results from the tasks (Tables \ref{tab:result-t1}-\ref{tab:result-t3}), it is evident that the proposed \textit{SCL} framework consistently converged in fewer epochs and required less training time, and thus demonstrating its advantage on computational efficiency.

Among the baselines, \textit{PTO-Sup} and \textit{Layer-LOF} also appeared to be efficient in terms of epoch count, but their training times were significantly longer. We consider the reason as \textit{PTO-Sup} required complete sentence generation and loss calculation for each word, leading to substantial asynchronous processing delays. \textit{Layer-LOF}, while not inherently inefficient, used adversarial attacks to oversample the data, which multiplied the training load. These factors inflated their total computation and resulted in suboptimal efficiency.

\subsection{Potential challenges, limitations and future solutions}\label{ssec:cha-lim}

While our method demonstrated strong performance across tasks, as a prototype, it still presents several challenges and limitations needed to be considered for practical deployment.

Background-shift in data: In T-2 and T-3, despite better results than \textit{FT}, the supervised nature of \textit{SCL} can still misalign OOD texts with background-shifts, drawing them closer to IND clusters. To address this, future work could include pre-evaluating background-shift intensity and minimizing pre-training epochs to balance IND intent classification and OOD detection accuracy. Incorporating unsupervised training targets may also help.

Noisy inputs: The filtering and re-training strategy between T-3 and T-4 could struggle with noisy inputs that resist categorization, complicating \textit{KMeans} clustering. The solution could be using advanced clustering methods like HDBSCAN, which can identify outliers without requiring a fixed cluster number. Alternatively, a filtering algorithm could be implemented to iteratively remove distant or scattered data points, though it risks accumulating noises over time.

Scalability on larger datasets: Task 4 results showed good scalability with the addition of up to 38 new intents. However, as category numbers increase, the model's ability to distinguish between highly similar classes may diminish. When this phenomenon is clearly observed, splitting the model into sub-models based on broader categories may be beneficial, and we plan to explore optimal splitting strategies in future work.

\section{Conclusion}
Traditional classification-retrieval-based QA systems primarily focus on categorizing known queries, which is insufficient to address the complexities of real-world scenarios. Moreover, the efficiency of the default training method is often suboptimal.
To address these issues, we introduced three new tasks — out-of-domain detection, new intent discovery, and continual learning — to enhance the system's functional capabilities. We then applied SCL as a unified representation learning method with minimal additional tuning to efficiently achieve the objectives of these tasks.
Experimental results demonstrated that, despite its simplicity, the proposed approach consistently outperformed state-of-the-art methods across all tasks.
These improvements, integrated within a natural workflow, significantly boost system adaptability and robustness, giving a meaningful impact on the future development of real-world QA systems.

While the proposed method is effective, it remains a prototype with room for further enhancement. 
Besides the challenges addressed in Section \ref{ssec:cha-lim}, future work could focus on integrating more advanced downstream methods to harness the full potential of SCL. For example, a larger-parameter encoder could provide a stronger representation base for learning, while more sophisticated analysis algorithms like energy-based methods should leverage SCL's tighter cluster formations for enhanced OOD detection effects. In addition, applying advanced techniques like DeepCluster \cite{caron2018deep} for new intent discovery and Elastic Weight Consolidation \cite{kirkpatrick2017overcoming} for continual learning could also further strengthen the system's performance.
Finally, the proposed SCL framework can also be extended to other QA tasks seamlessly, such as providing \textit{top-K} confidence results for ambiguous classifications or visualizing sub-intents through embedding space analysis.

\section*{Ethical and privacy statement}
In this study, the datasets used are all public-available and do not contain privacy-sensitive or confidential information. The neural networks used (i.e., BERT-series) are also less likely to inadvertently disclose sensitive information compared to generative models like GPTs.

\section*{Acknowledgment}
The authors would like to thank Seita Shimada and Yoshiaki Matsukawa, Rakuten Card Co. Ltd., for their helpful comments on this study.

\bibliographystyle{IEEEtran}
\bibliography{custom}
\newpage

\begin{IEEEbiography}[{\includegraphics[width=1in,height=1.25in,clip,keepaspectratio]{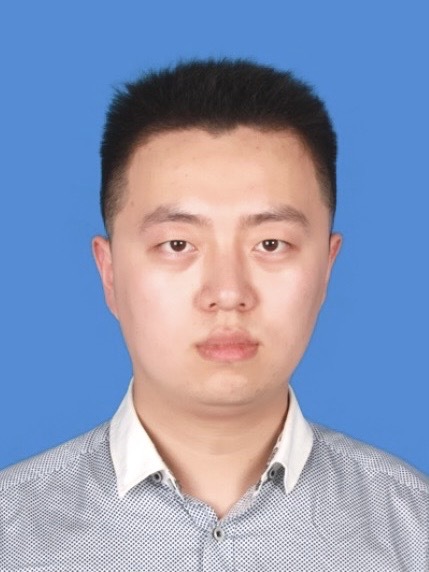}}]{Bo Wang} received the B.E. degree in computer science from Nanjing University of Posts and Telecommunications in 2018, and M.E. degrees in advanced information technology from Kyushu University in 2022. He is now pursuing the D.E. degree in information science and technology in Kyushu University.
His research interests focus on natural language processing, including user intent classification, out-of-domain text detection, and the general text representation learning. He has collaborated actively with company and institution researchers to conduct practical applications in question answering and automatic scoring systems.
\end{IEEEbiography}

\vspace{-13cm}
\begin{IEEEbiography}[{\includegraphics[width=1in,height=1.25in,clip,keepaspectratio]{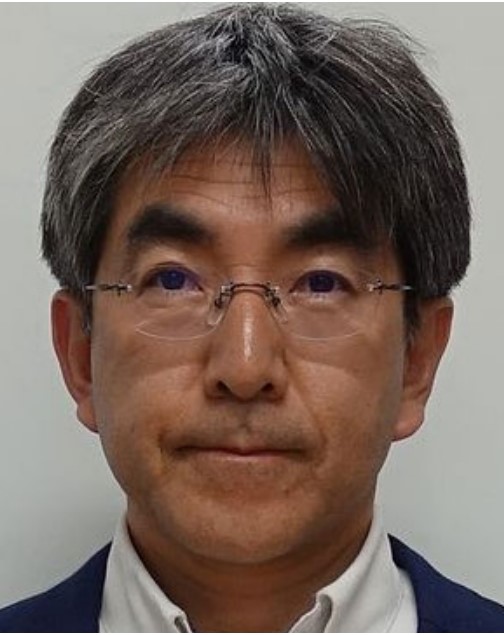}}]{Tsunenori Mine} (Member, IEEE) received the B.E. degree in computer science and computer engineering and the M.E. and D.E. degrees in information systems from Kyushu University, in 1987, 1989, and 1993, respectively. He is currently an Associate Professor with the Department of Advanced Information Technology, Faculty of Information Science and Electrical Engineering, Kyushu University. 
He is also leading several collaboration research projects with various companies and academic institutions to develop technologies and theories that are both practical and academically novel. His research interests include the development of real-world services using artificial intelligence techniques, especially natural language processing, text mining, data mining, recommendation, and multi-agent systems. He received the Best Paper Award from the Journal of the Information Processing Society of Japan (IPSJ) for his work on a parallel parsing algorithm, in 1993, and the IPSJ Activity Contribution Award, in 2014.
\end{IEEEbiography}

\newpage
\appendices
\begin{table*}[htbp]
	\centering
	\caption{Computation complexity of each method}
	\begin{tabular}{ll}
		\toprule
		& Total computation complexity $(O(model) + O(loss))$\\
		\midrule
		GENERAL METHOD &         \\
		\textit{CE}    & $O(model) + O(N * H * C)$  \\
		\multirow{2}[1]{*}{\textit{\textbf{SCL} (proposed)}} & $O(model) + O(N/N_b * N_b^2 * H)$ \\
		& $= O(model) + O(N * N_b * H)$  \\
		\midrule
		TASK 2 METHOD &       \\
		\textit{PTO-Sup (GPT-likelihood)} &  $O(model) + O(N*L^2*V) $  \\
		\textit{Layer-LOF (CE+layer redundancy under adv.)} & $K_{adv}*(O(model) + O(N*N_{layer}*H*C) + O(N*N_{layer}^2*H))$  \\
		\midrule
		TASK 3 METHOD &       \\
		\textit{MTP-KNNCL (mlm+knnConL)} & $O(model) + O(N*N_b*H*K_{knn})$  \\
		\textit{DPN (mlm+ce+prototype loss)} & $O(model) + O(N * H * C) + O(N * H * K_{proto})$  \\
		\midrule
		O(model) &   $O(N * (N_{layer} * H^2 + N_{layer}^2 * H))$    \\
		\bottomrule
	\end{tabular}%
	\label{tab:computation}%
\end{table*}%

\section{KMeans details}\label{appsec:kmeans}
The KMeans learning process is shown in Algorithm \ref{alg:kmeans}.
\begin{algorithm}[htbp]
	\caption{KMeans clustering}
	\newcommand{\Input}{\item[\textbf{Input:}]}
	\newcommand{\Output}{\item[\textbf{Output:}]}
	\begin{algorithmic}[1]
		\Input Text embeddings \(\{\boldsymbol{h}_{j'}\}_{j'=1}^{N_{ood}}\) of detected \(\{\boldsymbol{x}_{j'}\}_{j'=1}^{N_{ood}}\) in T-2, number of clusters \(K\)
		\Output Label assignments \(\{\hat{y}_{j'}\}_{j'=1}^{N_{ood}}\)
		
		\State Initialize \(K\) cluster centroids \(\{\boldsymbol{\mu}_k\}_{k=1}^K\) randomly.
		\Repeat
		\For{each data point \(\boldsymbol{h}_{j'}\)}
		\State Assign \(\boldsymbol{h}_{j'}\) to the nearest cluster centroid:
		\begin{equation}
			\small
			\hat{y}_{j'} = \arg\min_{k \in \{1, 2, \ldots, K\}} \|\boldsymbol{h}_{j'} - \boldsymbol{\mu}_k\|^2
		\end{equation}
		\EndFor
		
		\For{each cluster \(k\)}
		\State Update \(\boldsymbol{\mu}_k\) as the mean of all points assigned to cluster \(k\):
		\begin{equation}
			\small 
			\boldsymbol{\mu}_k = \frac{1}{|num(cate\ k)|} \sum \boldsymbol{h}_{j'\rightarrow k}
		\end{equation}
		\EndFor
		\Until{the cluster assignments \(\{\hat{y}_{j'}\}\) do not change or the centroids \(\{\boldsymbol{\mu}_k\}\) converge}
		
		\State \Return Label assignments \(\{\hat{y}_{j'}\}_{j'=1}^{N_{ood}}\)
	\end{algorithmic}\label{alg:kmeans}
\end{algorithm}

\begin{table}[ht]
	\centering
	\caption{Abbreviations and full names}
	\begin{tabular}{cl}
		\toprule
		Abbreviations & Full name \\
		\midrule
		QA    & question answering (system) \\
		RepL  & representation learning (feature learning)\\
		ConL  & contrastive learning \\
		MeL   & metric learning \\
		UConL & unsupervised contrastive learning \\
		SCL & supervised contrastive learning \\
		\midrule
		IND   & in-domain (texts)\\
		OOD   & out-of-domain (texts)\\
		\midrule
		GDA   & Gaussian discriminant analysis \\
		LOF   & local outlier factor \\
		\midrule
		GPT   & Generative Pre-Trained Transformer \\
		BERT  & Bidirectional Encoder Representations \\
		& from Transformers \\
		\midrule
		CE    & cross-entropy (loss) \\
		FT    & fine-tuning \\
		MLM   & masked language modeling \\
		NN   & nearest neighbor \\
		\bottomrule
	\end{tabular}%
	\label{tab:abbr}%
\end{table}%

\section{Metric calculation details} \label{appsec:metric}
Here we show the detailed calculation processes of the used metrics.
\subsection{For Task 1 and 4: F1 values}
The Micro and Macro F1 scores are calculated as follows:

\noindent\textbf{Micro F1 score}: is calculated by summarizing the classification situations over all classes. It is the harmonic mean of micro-averaged precision ($P$) and recall ($R$):

\begin{equation}
	\text{Micro F1} = \frac{2 \cdot P R}{P + R}
\end{equation}

\noindent\textbf{Macro F1 score}: calculates the F1 score independently for each class and then takes the average, which treats all classes equally:

\begin{equation}
	\text{Macro F1} = \frac{1}{C} \sum_{c=1}^{C} \text{F1-score}_c
\end{equation}
\noindent where $C$ is the number of categories.

\subsection{For Task 3: NMI, ARI, ACC}
The NMI, ARI, ACC are calculated as follows:

\noindent\textbf{Normalized Mutual Information (NMI)}: is a measure of the similarity between two sets of clusters:

\begin{equation}
	\text{NMI}(U, V) = \frac{2 \cdot I(U;V)}{H(U) + H(V)}
\end{equation}

\noindent where \(I(U;V)\) is the mutual information between clusters \(U\) and \(V\), and \(H(U)\) and \(H(V)\) are the entropies of \(U\) and \(V\) respectively. It ranges from 0 (no mutual information) to 1 (perfect correlation).

\noindent\textbf{Adjusted Rand Index (ARI)}: measures the similarity between two data clusterings by considering all pairs of samples, and counting pairs that are assigned in the same or different clusters in the predicted and true clusterings:

\begin{equation}
	\text{ARI} = \frac{\sum_{ij} \binom{n_{ij}}{2} - \left[ \sum_{i} \binom{a_{i}}{2} \sum_{j} \binom{b_{j}}{2} \right] / \binom{n}{2}}{\frac{1}{2} \left[ \sum_{i} \binom{a_{i}}{2} + \sum_{j} \binom{b_{j}}{2} \right] - \left[ \sum_{i} \binom{a_{i}}{2} \sum_{j} \binom{b_{j}}{2} \right] / \binom{n}{2}}
\end{equation}

\noindent where \(n_{ij}\) is the number of elements in the intersection of clusters \(i\) and \(j\), \(a_{i}\) is the sum over row \(i\), and \(b_{j}\) is the sum over column \(j\).

\noindent\textbf{Clustering Accuracy (ACC)}: evaluates the clustering result after finding the best one-to-one mapping between cluster labels and true labels:

\begin{equation}
	\text{ACC} = \frac{\sum_{i=1}^{n} \delta(y_i, \text{map}(\hat{y}_i))}{n}
\end{equation}

\noindent where \(y_i\) is the true label, \(\hat{y}_i\) is the cluster label, \(\text{map}()\) is the mapping function where we use the Hungary algorithm this time, and \(\delta\) is the Kronecker delta function.

\section{Computational complexity analysis}\label{app:eff}
The computational complexity analysis of each method is shown in Table \ref{tab:computation} according to the original papers.

Here, $N$ is the number of total data, $N_b$ is the number of data in one mini-batch and $N_{layer}$ is the number of layers in the network, where for bert-base it is 12. 
$H$ is the vector dimension of the hidden layer, for bert-base it is 768. $C$ is the number of IND categories. 
$L$ stands for the average input sentence length, while $V$ is the number of words in dictionary (specially for \textit{PTO-Sup}). Finally, $K_{adv}$ denotes the number of adversarial examples generated (specially for \textit{Layer-LOF}), $K_{knn}$ is the number of neighbors used in contrastive learning (specially for \textit{MTP-KNNCL}) and $K_{proto}$ is the number of assumed prototype categories (specially for \textit{DPN}).

\section{Learning and evaluation procedure (Algorithm form)}\label{appsec:data-usage}
The learning and evaluation procedures of pre-training and each task are shown in Algorithm \ref{alg:overall}.

\section{Abbreviations} 
Table \ref{tab:abbr} shows the abbreviations that have appeared in this paper and their complete names.

\begin{algorithm}[htbp]
	\caption{Overall process}
	\newcommand{\Input}{\item[\textbf{Input:}]}
	\newcommand{\Output}{\item[\textbf{Output:}]}
	\begin{algorithmic}[1]
		\Input Train data, Val data, Test I data, Test II data
		\Output Solutions for T-1 to T-4
		
		\State \qquad\qquad\qquad \textit{\textbf{-Pre-training Stage-}}
		\State $Epoch \gets 0$, $AUROC_{best-val} \gets 0$
		\While {$Epoch<20$}
		\State Train \textit{ENC} on $\boldsymbol{x}_{\text{train}}$ using SCL loss object (eq. 8)
		\State Calculate OOD detection $AUROC_{val}$ on this epoch
		
		\If {$AUROC_{val}>AUROC_{best-val}$}
		\State $AUROC_{best-val} = AUROC_{val}$
		\State Save model of this epoch, and record the threshold $\lambda$ when TPR=90\%
		\EndIf
		\State $Epoch = Epoch + 1$
		\EndWhile
		
		\State \qquad\qquad\qquad \qquad \textit{\textbf{-Task 1-}}
		\State Input $\boldsymbol{x}_{\text{test I}}$ (IND part) to \textit{ENC} to obtain the text representations:
		\State \qquad \qquad $\boldsymbol{h}_{\text{test I}} = ENC(\boldsymbol{x}_{\text{test I}})$
		\State The label is determined as the label of the nearest IND cluster centroid:
		\State \qquad \qquad $\hat{y}_{\text{test I}} = \arg\min_{c \in \mathcal{C}_{kn}} MDist(\boldsymbol{h}_{\text{test I}}, \bar{\boldsymbol{h}}_c)$
		\State Calculate Micro, Macro F1 values

		\State \qquad\qquad\qquad \qquad \textit{\textbf{-Task 2-}}
		\State Input $\boldsymbol{x}_{\text{test I}}$ (IND, OOD part) to \textit{ENC} to obtain the representations and calculate the detection scores:
		\State \qquad \qquad $sco(\boldsymbol x_{\text{test I}}) = {\underset{c\in {\mathcal C}_{kn}}{min}} \ MDist(\boldsymbol{h}_{\text{test I}}, \bar{\boldsymbol{h}}_c)$
		\State Calculate FPR, TPR results on various thresholds $\lambda$, and finally AUROC, AUPR, FPR90 values

		\State \qquad\qquad\qquad \textit{\textbf{-Task 3 preparation-}}
		\State Determine $\lambda_{filter}$ when OOD detection TPR=90\% on $\boldsymbol{x}_{\text{val}}$
		\State Divide $\boldsymbol{x}_{\text{test I}}$ into $\boldsymbol{x}_{\text{test I, `IND'}}$ and $\boldsymbol{x}_{\text{test I, `OOD'}}$ two parts by comparing detection $sco$ and $\lambda_{filter}$

		\State \qquad\qquad\qquad \qquad \textit{\textbf{-Task 3-}}
		\State Apply KMeans on $\boldsymbol{x}_{\text{test I, `OOD'}}$ and give new pseudo labels to $\boldsymbol{x}_{\text{test I, `OOD'}}$ as $\hat{y}_{\text{test I, `OOD'}} \in \mathcal{C}_{kn'}$
		\State Calculate NMI, ARI, ACC metrics

		\State \qquad\qquad\qquad \qquad \textit{\textbf{-Task 4-}}
		\State Give pseudo labels to $\boldsymbol{x}_{\text{test I, `IND'}}$ using the initial \textit{ENC} as $\hat{y}_{\text{test I, `IND'}} \in \mathcal{C}_{kn}$
		\State Train \textit{ENC} on $\{\boldsymbol{x}_{\text{test I}}, \hat{y}_{\text{test I}}\}$, $\hat{y}_{\text{test I}}$$ \in $$\{\mathcal{C}_{kn}$$+$$\mathcal{C}_{kn'}\}$ using SCL loss (eq. 8)
		\State Test on $\boldsymbol{x}_{\text{test II}}$ and calculate Micro, Macro F1 values
		\State \textbf{END}
	\end{algorithmic}
	\label{alg:overall}
\end{algorithm}

\EOD

\end{document}